\documentclass{article}
\PassOptionsToPackage{numbers, compress}{natbib}
\usepackage[preprint]{neurips_2026}

\usepackage[utf8]{inputenc} 
\usepackage[T1]{fontenc}    
\usepackage{hyperref}       
\usepackage{url}            
\usepackage{booktabs}       
\usepackage{amsfonts}       
\usepackage{nicefrac}       
\usepackage{microtype}      
\usepackage{xcolor}         
\usepackage{graphicx}
\usepackage{multirow}

\definecolor{darkgreen}{rgb}{0,0.5,0}

\title{EgoInteract: Synthetic Egocentric Videos Generation for Interaction Understanding and Anticipation}

\author{%
  Rosario Leonardi\textsuperscript{1,2}\thanks{Corresponding author: \texttt{rosario.leonardi@unict.it}} \quad
  Francesco Ragusa\textsuperscript{1,2} \quad
  Daniele Materia\textsuperscript{1} \\
  \textbf{Alessandro Passanisi}\textsuperscript{1} \quad
  \textbf{James Fort}\textsuperscript{3} \quad
  \textbf{Jakob Engel}\textsuperscript{3} \quad
  \textbf{Giovanni Maria Farinella}\textsuperscript{1,2} \\
  \vspace{0.3cm} \\
  \textsuperscript{1}Department of Mathematics and Computer Science, University of Catania, Italy \\
  \textsuperscript{2}Next Vision s.r.l., Catania, Italy \\
  \textsuperscript{3}Reality Labs Research, Meta, USA \\
}

\newcommand{\modelname}{\textbf{EgoInteract}}
\begin{document}

\maketitle

\begin{abstract}
Collecting large-scale egocentric video datasets with dense spatial and temporal annotations is costly, slow, and often constrained by environmental biases, privacy constraints, and limited coverage of interaction patterns. While synthetic data has shown strong potential in several vision domains, its use for egocentric perception remains relatively underexplored, especially for tasks requiring temporally coherent human-object interactions. In this work, we introduce \modelname{}, a controllable simulator for egocentric video generation designed to model fine-grained egocentric interactions and their temporal dynamics. The simulator enables precise control over camera, human body and hand motion, object manipulation, and scene composition across diverse environments. Building on this framework, we generate a synthetic egocentric video dataset with dense spatial and temporal annotations for temporal action segmentation, next-active object detection, interaction anticipation, and hand-object interaction detection. We evaluate models trained with simulated data on multiple real-world egocentric benchmarks spanning diverse environments, object categories, and interaction patterns. Results show consistent improvements over strong baselines across tasks and datasets, demonstrating the effectiveness and transferability of our simulation-based approach.
\end{abstract}

\section{Introduction}\label{sec_introduction}
Collecting large-scale egocentric video datasets with dense temporal and semantic annotations is notoriously time-consuming and expensive. Real-world recordings are subject to strong biases arising from the recording environment, the subject’s behavior, and safety or privacy constraints. Moreover, rare but critical interaction patterns, such as fine-grained hand-object manipulations or long-horizon action transitions, are difficult to capture at scale. As a result, existing datasets \cite{damen2018scaling, sener2022assembly101, ragusa2026enigma} often provide limited coverage of interaction dynamics, hindering the generalization of models across tasks and scenarios.

\begin{figure}
    \centering
    \includegraphics[width=0.8\linewidth]{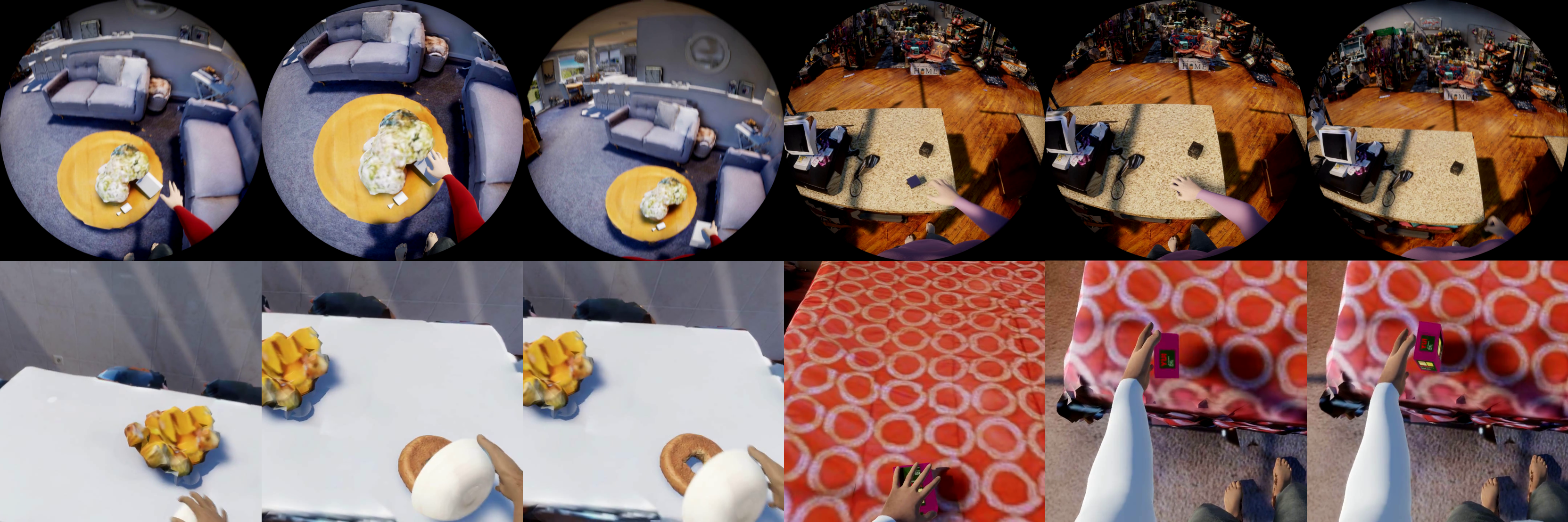}
    \caption{\modelname{} generates temporally coherent videos of humans interacting with diverse objects, enabling the study of egocentric interaction understanding at multiple levels.}
    \label{fig:examples_episodes}
\end{figure}

Data simulation offers a promising alternative to overcome these limitations. By generating synthetic data, simulators enable precise control over scene composition, object properties, camera motion, and interaction dynamics. Prior work has demonstrated that simulated data can substantially reduce the amount of real annotated data required to train models across several domains, including autonomous driving \cite{Dosovitskiy17, fabbri21iccv}, medical imaging \cite{Guo_2025_WACV, isaacSim_nvidia_2021}, and embodied AI \cite{ai2thor, szot2021habitat}.
Despite this success, while synthetic data has significantly advanced third-person vision models, its application to egocentric perception remains largely underexplored. Existing approaches have primarily focused on still images generation of hand-object interactions \cite{leonardi2025synthetic}, or on diffusion-based egocentric video synthesis methods that emphasize visual realism and world modeling rather than task-oriented supervision \cite{egocontrol, hao2026egosim}.

However, restricting simulation to static image limits the ability to model temporal phenomena that are intrinsic to egocentric perception \cite{li2024egogen}. Tasks related to egocentric interactions understanding, such as temporal action segmentation or interaction anticipation, require reasoning about motion, causality, and long-range temporal transitions, aspects that cannot be captured by isolated frames alone.

To address this challenge, we introduce \modelname{}, a controllable simulator designed for egocentric interaction understanding. Unlike prior synthetic data approaches limited to static image or unconstrained video generation, \modelname{} enables the generation of temporally coherent egocentric videos with explicit and fine-grained control over interaction dynamics (see Figure \ref{fig:examples_episodes}). The simulator allows precise settings of the egocentric camera behavior, human body and hand motions, and interactions with thousands of object categories across diverse environments. By explicitly modeling fine-grained hand-object interactions and their temporal evolution, \modelname{} provides a scalable and flexible platform for studying egocentric perception tasks that require reasoning over motion, causality, and procedural structure. 
Beyond data generation, \modelname{} provides rich and consistent annotations for the simulated egocentric videos, enabling supervised learning across multiple levels of egocentric human-object interactions tasks. The simulator automatically produces dense spatial annotations, including bounding boxes and semantic segmentation masks for both objects and hands as well as temporal annotations by assigning action labels with explicit start and end times, allowing the study of a wide range of egocentric tasks within a unified framework.

With the goal of studying egocentric human-object interactions at different levels, we conduct an extensive empirical evaluation demonstrating that simulated data generated with \modelname{} consistently improves performance across multiple real-world egocentric benchmarks. To this end, we generate a synthetic video dataset with dense spatial and temporal annotations. We leverage this dataset to study four representative egocentric tasks: 1) Temporal Action Segmentation, 2) Hand-Object Interaction Detection, 3) Next-Active Object Detection, and 4)  Interaction Anticipation. To demonstrate the strong generalization capabilities enabled by our simulator, we evaluate models trained with simulated data across multiple real-world egocentric benchmarks that differ substantially in environments, object categories, and interaction patterns, including EPIC-KITCHENS \cite{damen2018scaling}, HD-EPIC \cite{perrett2025hdepichighlydetailedegocentricvideo}, ENIGMA-51 \cite{ragusa2024enigma}, VISOR \cite{VISOR2022}, EgoHOS \cite{EgoHos_jianbo_eccv22}, Ego4D \cite{Grauman2021Ego4DAT}, Ego-Exo4D \cite{Grauman2025EgoExo4D}, and MECCANO \cite{ragusa_meccano_multimodal}. Overall, our findings suggest that \modelname{} can serve as a valuable complementary source of supervision for interaction-centric learning. Across multiple datasets and tasks, we observe consistent improvements when synthetic data is incorporated alongside real data. These observations motivate the development of flexible simulation frameworks and benchmarks that support systematic evaluation across both frame-based and temporal egocentric tasks. To foster future research, we publicly released the simulator, and the generated dataset at the following link: \url{https://fpv-iplab.github.io/EgoInteract/}.

\section{Related Work}\label{sec_related_work}

\subsection{Synthetic Data Simulation for Understanding Egocentric Interactions}

Most existing simulators focus on vehicles or robots in structured environments, supporting tasks such as autonomous driving and navigation \cite{Dosovitskiy17, xiazamirhe2018gibsonenv, habitat19iccv, isaacSim_nvidia_2021}. Advances in graphics and game engines have further enabled large-scale synthetic data generation for urban vision tasks \cite{rageEngine, fabbri21iccv}. More recent efforts have begun to address egocentric settings, including object-centric interaction modeling \cite{ai2thor} and geometry-focused data generation \cite{li2024egogen, avetisyan2024scenescript}. However, these approaches either lack explicit hand-object interaction modeling, focus on static imagery, or fail to capture temporal interaction dynamics. Closest to our work, recent pipelines for synthetic egocentric hand-object interaction data generation \cite{leonardi2025synthetic, leonardi2024exploiting} target HOI detection but remain limited to static images and constrained scenarios. To address these limitations, we propose \modelname{}, a simulator that explicitly models hand-object interactions and their temporal dynamics in egocentric video.

\subsection{Temporal Action Segmentation}
Temporal Action Segmentation (TAS) is crucial for egocentric interaction understanding, as it directly models the temporal evolution of human-object interactions in untrimmed videos. \\
The task has evolved from multi-stage pipelines with explicit temporal smoothing \cite{furnari2018personal} to end-to-end temporal models based on TCNs \cite{li2020ms, singhania2021coarse} and, more recently, Transformer-based architectures \cite{chinayi_ASformer, lu2024fact}. In this work, we use MS‑TCN++ \cite{li2020ms} as a representative baseline to study the effect of synthetic video augmentation.

\subsection{Egocentric Hand-Object Interaction Detection}
The task of \textit{Hand-Object Interaction} detection was firstly formulated as a combination of hand detection, contact state recognition, and object identification \cite{Shan2020UnderstandingHH}.
Prior work on egocentric hand-object interaction (HOI) understanding spans diverse task formulations and datasets, making direct comparison across methods challenging \cite{Ragusa2021TheMD, cheng2023towards, leonardi2022egocentric, ragusa2024enigma}. To address this, recent efforts have converged on the Hand-Object Segmentation (HOS) formulation adopted by benchmarks such as VISOR \cite{VISOR2022} and EgoHOS \cite{VISOR2022, EgoHos_jianbo_eccv22}. In this work, we follow the HOS setup to evaluate the effect of synthetic egocentric data generated by our simulator on VISOR, Ego4D \cite{Grauman2021Ego4DAT}, and ENIGMA-51 \cite{ragusa2024enigma}.

\subsection{Object Interaction Anticipation}
Anticipating future interactions is a fundamental component of egocentric interaction understanding, complementing the recognition of ongoing hand-object interactions.

Early studies on Next-Active Object anticipation leveraged motion and multimodal cues, such as object displacement, gaze, affordances, and hand trajectories, to predict imminent interactions \cite{furnari2017nextactiveobject, zhang2017deepfuturegaze, nagarajan2019grounded, liu2020forecasting}. More recent benchmarks have reframed anticipation as a high-level reasoning problem, including Ego4D’s short-term object interaction anticipation \cite{Grauman2021Ego4DAT} and HD-EPIC's VQA-based interaction anticipation for vision-language models \cite{perrett2025hdepichighlydetailedegocentricvideo}. State-of-the-art approaches further exploit structured visual cues such as gaze and segmentation \cite{materia2026leveraging}. In this work, we study whether synthetic egocentric data can improve interaction anticipation performance on the same benchmark.

\section{The \modelname{} Simulator}\label{sec_simulator}
We introduce \modelname{}, a Unity-based simulator for the generation of egocentric interaction data. \modelname{} enables the generation of first-person hand-object interaction episodes within diverse 3D environments, providing fine-grained control over agents, objects, camera behavior, and interaction parameters. Due to this high level of customization, the simulator supports a wide range of egocentric vision tasks, spanning both frame-based perception problems and temporally grounded interaction understanding. While in this work we focus on tasks to study interactions at multiple levels of granularity, \modelname{} is designed as a general simulation framework that can support many other egocentric vision tasks.

\subsection{Overview and Episode Definition}
\begin{figure}
    \centering
    \includegraphics[width=\linewidth]{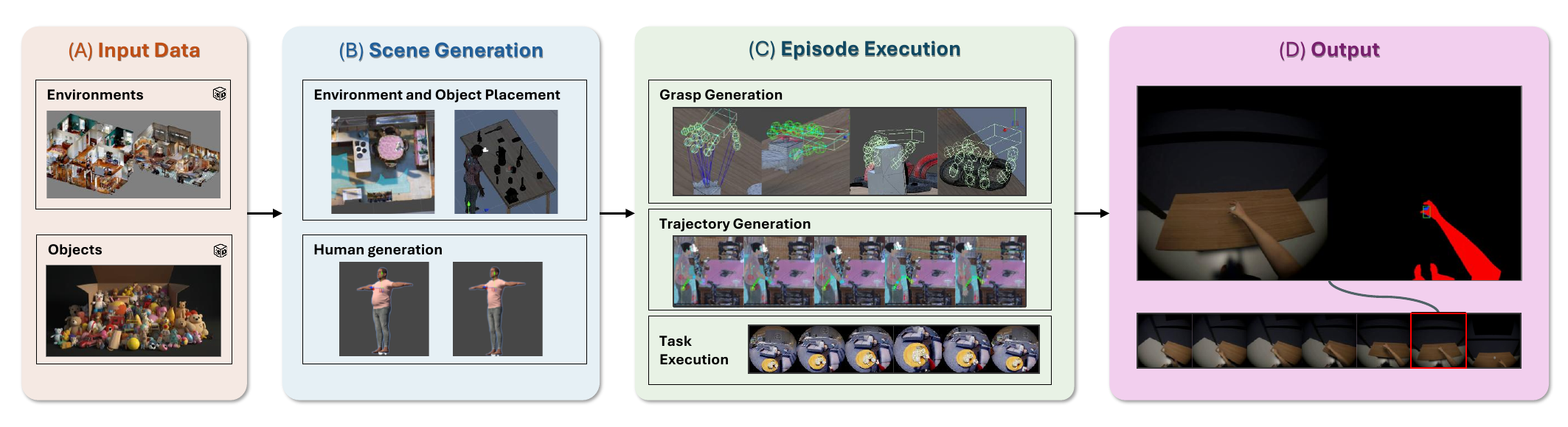}
    \caption{The \modelname{} simulator. The generated egocentric videos are automatically labeled with several spatial and temporal annotations.}
    \label{fig:main}
\end{figure}

In \modelname{}, interactions are organized as episodes, where each episode corresponds to a complete first-person interaction sequence centered on a single target object (Figure \ref{fig:main}). At the beginning of an episode, the agent is initialized in a rest pose near the target object and oriented toward it, while the interacting hand is sampled randomly. The agent then performs a grasping action on the target object, returns to a rest pose while holding the object, and finally executes a release phase in which the object is repositioned at a random new location in the environment. Additional distractor objects may also be present in the scene to increase environmental complexity. To increase the diversity of generated episodes, \modelname{} relies on a set of controlled randomization modules that vary scene layout, object placement, agent configuration, and interaction initialization and execution conditions across episodes.

This episodic formulation defines a temporally structured interaction sequence, making it suitable for supporting multiple egocentric prediction tasks from the same simulated sequence, ranging from frame-based perception to temporal interaction understanding. Although we focus on this episode structure, \modelname{} is flexible and can be extended to generate alternative interaction patterns and episode definitions.

\subsection{Input Data}
The first module is responsible of managing input environments and objects (Figure \ref{fig:main}-a).
For each episode, \modelname{} first samples an environment from a pool of 1,000 real-world indoor scenes from HM3D~\cite{habitat19iccv}. We then generate a multi-floor navigation surface over the scene geometry to identify the walkable areas available to the humanoid agents. To better capture narrow indoor passages, we configure the Unity NavMesh bake 
ensuring that the resulting walkable regions are compatible with the agent's physical dimensions and motion constraints.

Once the navigable structure of the environment is defined, \modelname{} populates the scene with the objects required for the sampled interaction episode. Target and distractor objects are selected from Objaverse XL\cite{objaverseXL}, a large-scale repository of diverse 3D assets covering a wide range of everyday object categories. In addition to these objects, \modelname{} supports the integration of custom 3D object assets with associated physical and semantic properties, allowing the simulator to be easily adapted to domain-specific scenarios without changes to the core simulation pipeline.

\subsection{Scene Generation}
The second module of \modelname{} is the Scene Generation (Figure \ref{fig:main}-b). Candidate placement locations are sampled from valid support surfaces within the environment, such as tabletops, shelves, and counters.
To identify such surfaces in a scene-agnostic manner, we exploit both the environment geometry and the multi-floor NavMesh generated for the humanoid agents. We cluster navigable regions according to their height, obtaining a set of floor-specific reference levels.
We validate object placement by generating an auxiliary NavMesh with a compact proxy agent (radius and height $0.1$), approximating average object size. This agent identifies feasible support regions, from which object locations are sampled while avoiding collisions with the environment and previously placed assets.

When the scene is fully initialized, \modelname{} instantiates a fully articulated embodied agent within the environment. Rather than relying on isolated floating hands, we model the agent using the SMPL‑X body model \cite{SMPL-X:2019}, which provides a coherent full-body representation that includes articulated hands and head pose~\footnote{We build on the Unity implementation available at: \url{https://gitlab.tuebingen.mpg.de/jtesch/smplx-unity}}. To animate the embodied agent, we rely on a full-body inverse kinematics system. A custom Unity-based IK~\footnote{Our implementation builds on the Final IK package available at: \url{https://assetstore.unity.com/packages/tools/animation/final-ik-14290?srsltid=AfmBOoqQ6RuOsyRMRh-HiaCKBJbQeQm_ROQ8xMB9f_eZpNLcxqwosvln}} solver drives the agent toward target poses defined by the sampled interaction, ensuring coherent full-body motion throughout the different interaction phases, including reaching, contact, grasp, and release. To increase agent visual diversity, we script variations in skin appearance and clothing. From the SMPL-X UV layout, we generate an alternative base texture offline using a vision-language model\footnote{We use Gemini for offline texture generation.}, then procedurally refine it in Unity to obtain diverse skin tones and clothing across episodes. At the beginning of each episode, we also sample different SMPL-X shape parameters to generate agents with diverse anthropometric characteristics, such as height-related proportions and overall body build.

Visual observations generated by the simulator are rendered from an egocentric camera rigidly attached to the agent’s head. The camera is positioned approximately at eye level and naturally follows the agent’s full‑body motion and posture, closely mimicking the viewpoint of wearable recording devices commonly used in real‑world egocentric datasets. In addition to this default configuration, \modelname{} supports extensive camera customization, allowing control over intrinsic and extrinsic parameters such as field of view, resolution, and relative camera offset. This flexibility enables the simulation of different wearable camera setups and supports adaptation to a wide range of egocentric vision scenarios. Together, these design choices increase both the visual and anthropometric diversity of the embodied agents while preserving realistic first-person interaction dynamics.

Additional implementation details are provided in the the technical appendix and supp. material.

\subsection{Episode Execution}

Given a target object, \modelname{} generates grasp configurations using a collider-based procedure. Object geometry is approximated with collision meshes generated via VHACD~\cite{mamou2016volumetric}, while the hand is represented using a simplified collision model with capsule colliders for the fingers and a box collider for the palm. A collision-aware hand proxy is iteratively advanced toward the object until initial palm contact is detected, defining a pre-grasp configuration. From this state, all fingers are simultaneously closed to form a power grasp, with collision checks active to reject invalid configurations. A grasp is accepted only if a geometric opposition metric indicates a valid thumb-finger enclosure of the object. The procedure is repeated for up to 50 trials, retaining the first valid grasp.
Once a grasp is obtained, \modelname{} plans interaction trajectories using Bézier curves between the initial hand pose and the grasp target. Candidate trajectories are validated through collision checking using a kinematic agent proxy, and only collision-free trajectories are executed via the full-body inverse kinematics system. After grasping, the object is attached to the hand, the agent returns to a rest pose, and a release phase is performed by sampling a reachable placement location and validating a corresponding placement trajectory following the same collision-aware procedure.

\subsection{Multi-Task Annotations and Outputs}
To support multi-task evaluation, \modelname{} automatically extracts several dense annotations that are difficult and expensive to acquire in real-world egocentric video. To meet the annotation requirements of our benchmark, we extended the Unity Perception package~\footnote{\url{https://docs.unity3d.com/Packages/com.unity.perception@1.0/manual/index.html}} with custom labelers and metadata exporters tailored to egocentric hand-object interactions and anticipation tasks.

In addition to rendering egocentric visual observations, \modelname{} automatically generates rich and structured annotations aligned with the underlying interaction dynamics. On the visual side, the simulator outputs egocentric RGB frames, depth maps, and instance‑level segmentation masks. Beyond raw sensory data, \modelname{} provides a comprehensive set of semantic and temporal annotations that describe how interactions evolve over time.
Each interaction episode is annotated with frame-level action labels and precise timestamps for key interaction events, enabling a fine-grained temporal decomposition of the interaction sequence. The simulator further encodes detailed hand-object relationship annotations, including hand assignments, contact states, and explicit hand-object associations that evolve throughout the episode. Object-centric metadata such as object identity and state changes are also available, together with spatial annotations in the form of bounding boxes and segmentation masks.
This annotation design makes \modelname{} applicable to a wide range of egocentric perception tasks, spanning both frame-based and temporal settings. Figure~\ref{fig:main}-d shows examples of the visual streams and task-specific annotations generated by \modelname{}.

\subsection{The \modelname{} Dataset}
The \modelname{} dataset is designed as a benchmark for multi-task egocentric interaction understanding. It consists of temporally structured episodes generated by the simulator, resulting in diverse interaction instances with consistent annotations across tasks.
Each episode is automatically annotated with rich temporal annotations, including action labels and precise start and end times, as well as spatial annotations, such as bounding boxes and semantic masks for hands and active objects, and explicit hand-object relations. In addition, the dataset includes future-oriented annotations supporting anticipation tasks, such as next‑active object bounding boxes and VQA interaction questions with one correct answer and four carefully sampled distractors. 
In total, the dataset contains 10,534 generated episodes recorded at 30 FPS, corresponding to approximately 1.9M frames overall. 
Each episode lasts about 6 seconds on average. Since the generated data are used exclusively for training, no train/validation/test split is defined.

Together, these resources make the dataset suitable for both frame-level and temporal tasks. The released data are provided in both Aria and GoPro formats, allowing the benchmark to reflect different egocentric capture configurations. Examples of the generated episodes are in the supp. material.

\section{Experiments and Results}\label{sec_experimental_setup}
To systematically study egocentric human-object interactions at different levels, we used the \modelname{} dataset. We leverage this dataset to evaluate the impact of simulated training data on four representative egocentric tasks. Additional details and qualitative results are reported in the supp. material. 

\subsection{Temporal Action Segmentation} 
Temporal Action Segmentation aims to recognize and segment long, untrimmed videos into a sequence of semantically meaningful actions by assigning an action label, to each frame or temporal segment while accurately localizing action boundaries. We adopt an interaction-centric formulation of the task by focusing exclusively on the \textit{Take} and \textit{Release} actions. These actions represent fundamental transitions in object manipulation, marking when an object becomes engaged in or disengaged from an interaction, and therefore play a critical role in understanding the temporal dynamics of human-object interactions in egocentric video.
\paragraph{Datasets.}
\textbf{EPIC-Kitchens-100 \cite{Damen2021RESCALING}:} consists of over 100 hours of egocentric video capturing daily activities in kitchen environments and includes 89,977 temporally annotated action segments. In this work, we focus exclusively on \textit{Take} and \textit{Release} actions, grouping together verb classes that are semantically aligned with object acquisition (e.g., \textit{take cup}, \textit{take plate}, \textit{take carrot}) and object disengagement (e.g., \textit{put-down plate}, \textit{put-down fork}, \textit{put-on container}). Under this mapping, the dataset contains a total of 13,091 \textit{Take} segments and 10,873 \textit{Release} segments. 
\textbf{Ego-Exo4D \cite{Grauman2025EgoExo4D}:} is a large-scale multimodal dataset capturing human activities from both egocentric and exocentric viewpoints across diverse environments. The dataset includes 143,442 temporally annotated segments spanning 664 keysteps across 17 high-level activities. In our experiments, we consider only the egocentric subset and adapt the provided annotations to our interaction-centric setting by mapping object retrieval keysteps to \textit{Take} and object storage keysteps to \textit{Release}. Specifically, we identified 82 keystep classes corresponding to \textit{Take} actions (e.g., \textit{Get knife}, \textit{Get a pot or saucepan}) and 81 keystep classes corresponding to \textit{Release} actions (e.g., \textit{Put away plate}, \textit{Put away spatula}), resulting in 1,778 \textit{Take} segments and 480 \textit{Release} segments.\\
 
\textbf{Settings.}
We evaluate the considered baseline using varying proportions of labeled real training data (10\%, 25\%, 50\%, and 100\%) to analyze the effect of synthetic data augmentation across different levels of real-data availability. Due to the substantial visual and distributional gap between synthetic data and real-world egocentric videos in EPIC‑Kitchens, we adopt a Domain‑Adversarial Neural Network (DANN) training strategy \cite{JMLR:v17:15-239}. This approach explicitly encourages the learning of domain-invariant representations, mitigating appearance differences while preserving task-relevant discriminative features.
In contrast, for Ego‑Exo4D, the visual characteristics of the egocentric data, such as camera placement, interaction scale, and scene structure, are more closely aligned with those of the simulated domain. As a result, the domain gap is comparatively smaller, and we observe that standard joint training with mixed real and synthetic samples is sufficient. For this dataset, we therefore follow a combined training protocol without adversarial domain adaptation.\\
\textbf{Baseline.} We adopt MS‑TCN++ \cite{li2020ms} as a representative baseline to evaluate whether augmenting real training data with synthetically generated egocentric videos consistently improves performance in Temporal Action Segmentation.\\
\textbf{Evaluation Measures.} We use standard metrics following prior work \cite{sener2022assembly101, lea2017temporal}, reporting F1 scores overlapping thresholds of 10\%, 25\%,and 50\%, denoted by F1@10,25,50. \\
\textbf{Results.}
Table \ref{tab:tas_results} reports the TAS results on Ego-Exo4D (left) and EPIC-KITCHENS (right). Incorporating synthetic data generated by \modelname{} generally improves upon the Real-only baseline, with the most substantial gains concentrated at the intermediate supervision level. At 10\% real data, synthetic samples fail to help on both datasets, suggesting that minimal real supervision is insufficient to effectively anchor the learning process. At 25\%, the improvement becomes substantial, with RS outperforming the Real-Only baseline by (+2.89 F1@10, +1.23 F1@25, +1.24 F1@50) on Ego-Exo4D and (+8.46 F1@10, +7.19 F1@25, and +4.19 F1@50) on EPIC-KITCHENS. At full supervision (100\%) the contribution of  synthetic data becomes marginal in both datasets, indicating that \modelname{} is most beneficial when real annotations are scarce.

\begin{table*}[t]
	\centering
	\caption{%
		Temporal action segmentation results across different proportions of labeled real training data.
		\textbf{R}\,=\,real data only; \textbf{RS}\,=\,real\,+\,synthetic.
		\textbf{Bold} indicates the best result within each R/RS pair.
	}
	\label{tab:tas_results}
	\setlength{\tabcolsep}{4pt}
	\resizebox{0.7\textwidth}{!}{%
		\begin{tabular}{l c |  r r r |  r r r }
			\toprule
			& & \multicolumn{3}{c|}{\textbf{Ego-Exo4D}} 
			& \multicolumn{3}{c}{\textbf{EPIC-KITCHENS}}  \\
			
			\textbf{Split} & \textbf{Config}
			& F1@10 & F1@25 & F1@50
			& F1@10 & F1@25 & F1@50 \\
			\midrule
			
			\multirow{2}{*}{10\%}
			  & R  & \textbf{35.18} & \textbf{29.80} & \textbf{15.51} 
			& 24.07 & \textbf{19.44} & \textbf{10.74} \\
			  & RS & 32.72          & 25.50          & 12.99          
			& \textbf{24.60} & 19.08 & 9.89 \\
			\midrule
			
			\multirow{2}{*}{25\%}
			  & R  & 37.73          & 31.37          & 17.70          
			& 24.49 & 19.96 & 11.46 \\
			  & RS & \textbf{40.62} & \textbf{32.60} & \textbf{18.94} 
			& \textbf{32.95} & \textbf{27.15} & \textbf{15.65} \\
			\midrule
			
			\multirow{2}{*}{50\%}
			  & R  & 41.61          & 34.90          & 22.44          
			& 33.91 & 28.25 & 17.62 \\
			  & RS & \textbf{43.01} & \textbf{35.59} & \textbf{22.65} 
			& \textbf{35.06} & \textbf{29.25} & \textbf{17.76}\\
			\midrule
			
			\multirow{2}{*}{100\%}
			  & R  & 40.97          & 34.17          & 21.29          
			& \textbf{41.86}  &    \textbf{35.73}   &     21.24    \\
			  & RS & \textbf{41.08} & \textbf{35.03} & \textbf{21.84} 
			& 40.36  & 35.02 &  \textbf{22.16}\\
			\bottomrule
		\end{tabular}%
	}
\end{table*}
\subsection{Egocentric Hand-Object Interaction Detection}
Modeling hand-object interactions in egocentric video is essential for interaction understanding, as it provides an explicit representation of the physical relationships between hands and objects. We adopt the HOS formulation \cite{VISOR2022}, which models egocentric hand-object interaction detection as the joint segmentation of hands and objects and the estimation of their contact relationships at the pixel level.

\paragraph{Datasets.} 
\textbf{VISOR \cite{VISOR2022}:} consists of 36 hours of egocentric video sampled from EPIC-KITCHENS-100 \cite{Damen2021RESCALING} and includes 32,857 training images with pixel-wise annotations covering 42,787 hand–object interaction instances. 
\textbf{EgoHOS \cite{EgoHos_jianbo_eccv22}:} contains 8,107 egocentric training images depicting hand–object interactions, sparsely sampled from videos in Ego4D \cite{Grauman2021Ego4DAT}, THU‑READ \cite{Tang2017ActionRI}, EPIC‑KITCHENS \cite{damen2018scaling}, and additional egocentric recordings from escape room scenarios. The dataset provides pixel-wise annotations for 13,659 hand-object relations. 
\textbf{ENIGMA-51 \cite{ragusa2024enigma}:} is an egocentric dataset capturing industrial activities in which participants follow procedural instructions to repair electrical boards. It comprises 51 videos totaling approximately 22 hours of footage and 45,505 labeled images. \\
\textbf{Settings.}
We evaluate hand-object interaction segmentation under different proportions of labeled real training data (0\%, 10\%, 25\%, 50\%, and 100\%). For experiments involving synthetic data generated by \modelname{}, we follow the training strategy of \cite{leonardi2025synthetic}, which builds on Adaptive Teacher \cite{li2022cross} to leverage labeled synthetic data together with real data under a domain adaptation setting. \\
\textbf{Baseline.} We adopt VISOR‑HOS \cite{VISOR2022} as the baseline method for egocentric hand-object interaction segmentation. VISOR-HOS builds upon the PointRend instance segmentation framework \cite{kirillov2020pointrend} and extends it with dedicated prediction heads to model interaction-specific attributes. 
This architecture provides a strong and widely used baseline for evaluating hand-object interaction understanding \cite{VISOR2022}.\\
\textbf{Evaluation Measures.} Following \cite{VISOR2022}, we evaluate hand-object interaction segmentation using the Hand + Object (Overall) mAP, which provides a unified evaluation of hand and object segmentation quality together with hand contact prediction and hand-object association accuracy. \\
\textbf{Results.}
Table \ref{tab:joint_results}-left reports the results on the VISOR, EgoHOS, and ENIGMA-51 datasets. Across all datasets, augmenting real training data with synthetic samples generated by \modelname{} consistently improves performance over the Real-Only baseline at all proportions of available real data. Notably, in the 50\% real-data setting, models trained with synthetic augmentation achieve higher performance on VISOR (46.26 vs. 45.33) and EgoHOS (39.94 vs. 36.16), and comparable performance on ENIGMA‑51 (63.03 vs. 63.84), relative to models trained using 100\% of the real data alone. These results highlight the effectiveness of synthetic data in substantially reducing the amount of real-world annotation required for egocentric hand-object interaction understanding.

\begin{table}[t]
	\centering
	\caption{Performance across egocentric interaction tasks.
		Left: Hand-Object Interaction results on VISOR, EgoHOS, and ENIGMA-51.
		Right: Next-Active Object detection on MECCANO and Ego4D (AP50:95 / AP50).
		\textbf{Bold} indicates best results.}
	\label{tab:joint_results}
	
	\setlength{\arrayrulewidth}{1.2pt} 
	
	\resizebox{0.9\linewidth}{!}{%
		\begin{tabular}{lc|ccc|cc}
			\toprule
			\multicolumn{2}{c|}{\textbf{Setting}} &
			\multicolumn{3}{c|}{\textbf{Hand-Object Interaction}} &
			\multicolumn{2}{c}{\textbf{Next-Active Object}} \\
			\cmidrule(lr){1-2}
			\cmidrule(lr){3-5}
			\cmidrule(lr){6-7}
			
			\textbf{Split} & \textbf{Config} &
			\textbf{VISOR} & \textbf{EgoHOS} & \textbf{ENIGMA-51} &                
			\textbf{MECCANO} & \textbf{Ego4D} \\
			\midrule
			
			0\% & Synth-Only 
			               & 30.65           & 20.18              & 24.35          
			& 02.82 / 06.30 & 00.51 / 01.22 \\
			
			\midrule
			\multirow{2}{*}{10\%} & Real-Only 
			               & 38.55           & 28.44              & 45.39          
			& 09.32 / 23.41 & 02.10 / 05.71 \\
			& Synth+Real 
			               & \textbf{41.83}  & \textbf{33.54}     & \textbf{46.46} 
			& \textbf{18.23} / \textbf{34.40} & \textbf{05.61} / \textbf{10.88} \\
			
			\midrule
			\multirow{2}{*}{25\%} & Real-Only 
			               & 37.90           & 33.73              & 51.83          
			& 13.22 / 30.93 & 03.41 / 08.39 \\
			& Synth+Real
			               & \textbf{43.75}  & \textbf{35.94}     & \textbf{57.32} 
			& \textbf{22.08} / \textbf{41.97} & \textbf{06.06} / \textbf{11.84} \\
			
			\midrule
			\multirow{2}{*}{50\%} & Real-Only 
			               & 38.15           & 36.30              & 57.62          
			& 15.56 / 34.31 & 04.95 / 10.43 \\
			& Synth+Real 
			               & \textbf{46.26}  & \textbf{39.94}     & \textbf{63.03} 
			& \textbf{23.46} / \textbf{44.34} & \textbf{07.11} / \textbf{13.46} \\
			
			\midrule
			\multirow{2}{*}{100\%} & Real-Only 
			               & 45.33           & 36.16              & 63.84          
			& 17.52 / 36.38 & 05.72 / 11.64 \\
			& Synth+Real 
			               & \textbf{46.20}  & \textbf{40.78}     & \textbf{66.07} 
			& \textbf{24.34} / \textbf{45.54} & \textbf{07.46} / \textbf{14.08} \\
			
			\bottomrule
		\end{tabular}}
\end{table}

\subsection{Next-Active Object Detection}
Beyond recognizing current hand-object interactions, egocentric understanding requires predicting future interactions, such as identifying the next object a user will actively engage with. We follow the formulation of Next-Active Object detection introduced in \cite{ragusa_meccano_multimodal}, which aims to anticipate objects involved in upcoming interactions.
In contrast to prior work, we adopt a class-agnostic formulation, focusing on predicting the instance of the next active object rather than its semantic category. This design choice allows us to emphasize interaction dynamics and object relevance independently of category recognition, and aligns naturally with scenarios involving previously unseen objects or long-tailed object distributions.

\paragraph{Datasets.}
\textbf{MECCANO \cite{ragusa_meccano_multimodal}:} is an egocentric dataset focusing on human-object interactions in an industrial assembly scenario. The dataset includes recordings from 20 subjects who assemble a toy motorbike, resulting in 20 egocentric video sequences with an average duration of 20.79 minutes each. MECCANO is widely used for studying interaction understanding and anticipation in structured industrial environments. 
\textbf{Ego4D \cite{Grauman2021Ego4DAT}:} is a large-scale egocentric dataset capturing daily-life activities across a wide range of unscripted scenarios, including household, outdoor, workplace, and leisure environments. It comprises 3,670 hours of egocentric video recorded by 931 unique camera wearers across 74 locations in 9 countries, making it one of the most extensive first-person video datasets.\\
\textbf{Settings.}
Following the same protocol adopted for the previous tasks, we evaluate the baseline using varying proportions of real data (0\%, 25\%, 50\%, and 100\%). In this setting, we employ the Adaptive Teacher \cite{li2022cross} approach to perform domain adaptation, reducing the gap between synthetic and real data. For the Ego4D dataset, the forecasting labels were adapted to match our formalized NAO task. \\
\textbf{Baseline.}  We adopt the baseline proposed in MECCANO \cite{ragusa_meccano_multimodal}, which is based on a Faster R-CNN detector with a ResNet-101 backbone. The model is trained to detect candidate objects in the current frame and to anticipate which object instance will become active in the future interaction. \\
\textbf{Evaluation Measures.} 
We adopt Average Precisions (APs) \cite{coco_dataset, Everingham10} as the evaluation metrics, assessing the accuracy of the predicted object instances solely based on spatial localization.\\
\textbf{Results.}
Table \ref{tab:joint_results}-right reports the results on the MECCANO and Ego4D datasets. Across both benchmarks and all experimental settings, incorporating synthetic data generated by \modelname{} consistently improves performance over training with real data alone. Notably, in the 25\% real-data setting, models augmented with synthetic data achieve performance that surpasses that obtained using 100\% of the real data, both on MECCANO (41.97 vs. 36.38) and Ego4D (11.84 vs. 11.64). These results indicate that synthetic augmentation can substantially reduce the amount of real annotated data required while improving anticipation performance.

\subsection{Interaction Anticipation}
Following \cite{perrett2025hdepichighlydetailedegocentricvideo}, we formulate interaction anticipation as a VQA task in which the model selects the next interacting object from a set of five candidates given a trimmed egocentric video clip.

\paragraph{Datasets.}
\textbf{HD-EPIC \cite{perrett2025hdepichighlydetailedegocentricvideo}:} contains 41 hours of unscripted egocentric kitchen videos with dense multimodal annotations. The anticipation benchmark consists of 1,000 multiple-choice questions paired with 10 second egocentric video clips, each ending shortly after gaze priming of the next interaction object. Each question includes one correct object and four distractors sampled from other objects manipulated within the video.\\
\textbf{Settings.}
We finetune two open-source VLLMs on synthetic video sequences generated by \modelname{}. We apply LoRA~\cite{hu2021loralowrankadaptationlarge} on all projection layers of the language backbone, keeping the vision encoder frozen, with $r{=}8$, $\alpha{=}16$, lr $5{\times}10^{-6}$, and 3 epochs.\\
\textbf{Baseline.} We adopt LLaVA-OneVision-7B~\cite{li2024llavaonevisioneasyvisualtask} and LLaVA-OneVision-1.5-8B-Instruct~\cite{an2025llavaonevision15fullyopenframework}. Their architectural differences allow us to assess whether improvements generalise across models. Following~\cite{materia2026leveraging}, each model is evaluated under four visual augmentation modes: \emph{Standard}, \emph{SoM}, \emph{Gaze}, and \emph{SoM\,+\,Gaze}, at $n{=}15$ frames with uniform sampling ($\lambda{=}0$).\\
\textbf{Evaluation Measures.} 
Since the VQA task is formulated as a multiple-choice prediction problem, we adopt standard classification accuracy as the evaluation metric.\\

\textbf{Results.}
Table~\ref{tab:oia_15} shows that synthetic finetuning improves upon the original baselines across most evaluation modes. For LLaVA-OV-7B, the largest gains are observed when gaze information is present in the input, reaching $+1.9$\% accuracy on SoM+Gaze and $+0.7$\% on Gaze; Standard and SoM modes instead show a marginal degradation ($-0.3$\%). For LLaVA-OV-1.5-8B, finetuning consistently improves over the original baseline across all modes, with gains ranging from $+0.1$\% on Standard up to $+1.0$\% on SoM+Gaze. Overall, these results demonstrate that synthetic data can meaningfully enhance interaction anticipation performance even for VLLMs trained on vast real-world corpora, providing measurable gains without requiring additional task-specific real annotations. This highlights the complementary role of simulation-based supervision given by \modelname{} in refining high-level reasoning capabilities beyond large-scale pretraining alone.

\begin{table}[t]
    \centering
    \caption{Accuracy (\%) on the \emph{HD-EPIC Interaction Anticipation} benchmark.}
    \label{tab:oia_15}
    \resizebox{0.6\linewidth}{!}{%
    \begin{tabular}{lcccccccc}
    \toprule
    & \multicolumn{2}{c}{\textbf{LLaVA-OV-7B}} & & \multicolumn{2}{c}{\textbf{LLaVA-OV-1.5-8B}} \\
    \cmidrule{2-3} \cmidrule{5-6}
    \textbf{Mode} & \textbf{Original} & \textbf{Finetuned} & & \textbf{Original} & \textbf{Finetuned} \\
    \midrule
    Standard & \textbf{18.3} & 18.0\,{\scriptsize($-$0.3)} & & 19.8 & \textbf{19.9}\,{\scriptsize($+$0.1)} \\
    SoM      & \textbf{18.7} & 18.4\,{\scriptsize($-$0.3)} & & 16.9 & \textbf{17.6}\,{\scriptsize($+$0.7)} \\
    Gaze     & 20.2 & \textbf{20.9}\,{\scriptsize($+$0.7)} & & 17.8 & \textbf{18.7}\,{\scriptsize($+$0.9)} \\
    SoM+Gaze & 20.0 & \textbf{21.9}\,{\scriptsize($+$1.9)} & & 15.1 & \textbf{16.1}\,{\scriptsize($+$1.0)} \\
    \bottomrule
    \end{tabular}}
\end{table}

\section{Limitations and Conclusions}\label{sec_conclusion}
In this work, we introduced \modelname{}, a highly controllable simulator for generating egocentric interaction data. We leveraged \modelname{} to build synthetic benchmarks and demonstrate consistent performance gains across multiple egocentric tasks. Our results show that synthetic data can substantially reduce the reliance on large amounts of real annotated data while maintaining strong performance. Despite these results and the strong generalization observed across multiple benchmarks, the simulator currently focuses on single-agent interactions involving a single target object and does not model social or collaborative scenarios. Addressing these limitations, by supporting more complex interaction dynamics involving multiple objects and multi-agent settings, represents a promising direction for future work. We believe \modelname{} and the associated benchmarks provide a flexible foundation for future research in egocentric vision and interaction understanding.

\begin{ack}
Research at University of Catania has been supported by Meta, Next Vision, and by the project Future Artificial Intelligence Research (FAIR) – PNRR MUR Cod. PE0000013 - CUP: E63C22001940006.
\end{ack}

\bibliographystyle{plainnat}
\bibliography{bibliography.bib}

\newpage

\appendix

\section{Technical appendices}
This supplementary material provides additional technical details, implementation specifics, ablation results, visual examples, and qualitative results of our study that complement the main paper.

\subsection{Additional Details on the \modelname{} Simulator}
\begin{figure}
    \centering
    \includegraphics[width=0.75\linewidth]{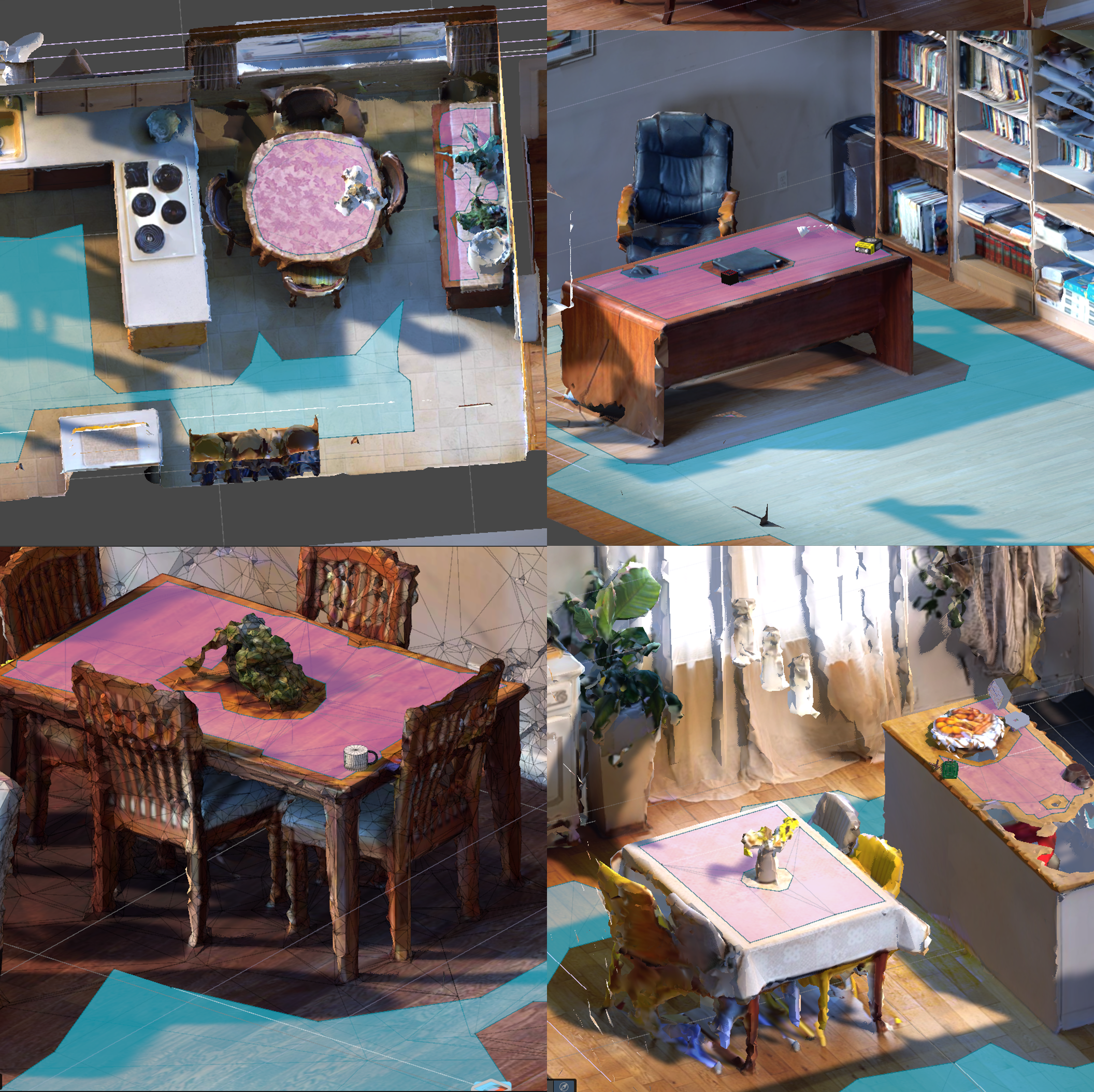}
    \caption{Examples of HM3D environments used in \modelname{}. Blue regions denote the navigable surfaces used for agent placement and motion, while purple regions indicate support surfaces where objects can be placed.}
    \label{fig:navmesh}
\end{figure}

\paragraph{Environment Initialization}
Each episode begins by sampling a 3D environment from the HM3D dataset. For each sampled scene, we extract both the navigation mesh used for valid agent placement and motion, and the support surfaces used for object placement, as shown in Fig.~\ref{fig:navmesh}.

To increase variability across episodes while preserving the scene structure, \modelname{} applies randomization modules during environment initialization. In particular, we randomize scene illumination through a \textit{sun-angle randomizer} that samples the hour of day uniformly between 8 and 16, the day of the year over the full annual range, and the latitude uniformly between $-45^\circ$ and $45^\circ$, producing diverse lighting conditions within the same environment.

The sampled environment is kept fixed for \texttt{10} consecutive episodes before a new HM3D scene is selected. At each episode reset, the randomization modules are re-applied so that different episodes may share the same environment while still exhibiting different visual conditions. To better capture narrow indoor passages, we configure the Unity NavMesh bake with a voxel size of $0.8$, a tile size of $256$, a minimum region area of $2$, and height-mesh generation enabled. The navigation surface is generated for a humanoid agent with radius $0.25$ and height $2$, ensuring that the resulting walkable regions are compatible with the agent's physical dimensions and motion constraints.
For each floor with height \(h_f\), we then identify candidate support regions by selecting scene geometry centered around \(h_f + 0.8\) m within a tolerance range of $0.4$ m. 

\paragraph{Object Initialization}
Given a sampled environment, each episode is populated with objects drawn from a fixed catalog derived from Objaverse-XL. In our current setup, this catalog contains 148,385 candidate assets, which are used to instantiate both target and distractor objects. Since Objaverse-XL assets exhibit substantial variability in scale, we preprocess object meshes by rescaling them following the strategy adopted in Grasp-XL. This normalization step yields object dimensions that are more suitable for grasp generation and interaction execution within the simulator.

\begin{figure}
    \centering
    \includegraphics[width=1\linewidth]{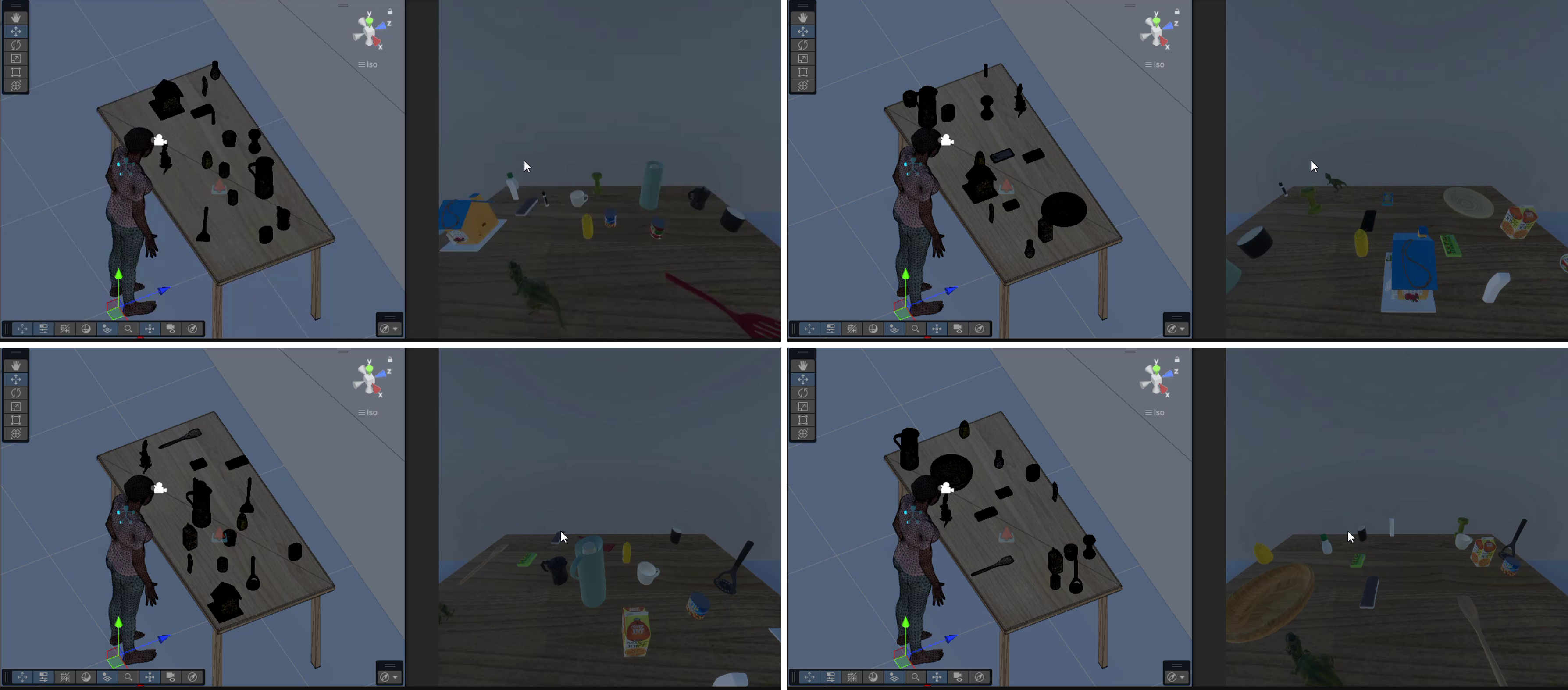}
    \caption{Examples of procedurally generated tabletop scenes in \modelname{}. For each example, the left image shows the global scene configuration, while the right image shows the corresponding egocentric view observed by the embodied agent.}
    \label{fig:obj_place}
\end{figure}

Object placement is performed by sampling valid support regions under a non-overlap constraint and rejecting configurations that intersect the environment or previously placed assets. This yields cluttered yet valid initial scenes while preserving procedural diversity across episodes. Examples of the resulting scene layouts and corresponding egocentric observations are shown in Fig.~\ref{fig:obj_place}.

A target object is then selected uniformly at random from the subset of instantiated objects that are reachable by the agent. Once the target is defined, the agent is positioned at the closest valid location to the target on the navigation mesh and oriented toward it, ensuring that each episode starts from a feasible embodied configuration.

\paragraph{Agent Initialization}
Once the target object has been defined, the embodied agent is instantiated in the scene and positioned at the closest valid point to the target on the navigation mesh. The agent is then oriented toward the target object, ensuring that each episode starts from a feasible embodied configuration compatible with the scene geometry and navigation constraints.

\begin{figure}
    \centering
    \includegraphics[width=1\linewidth]{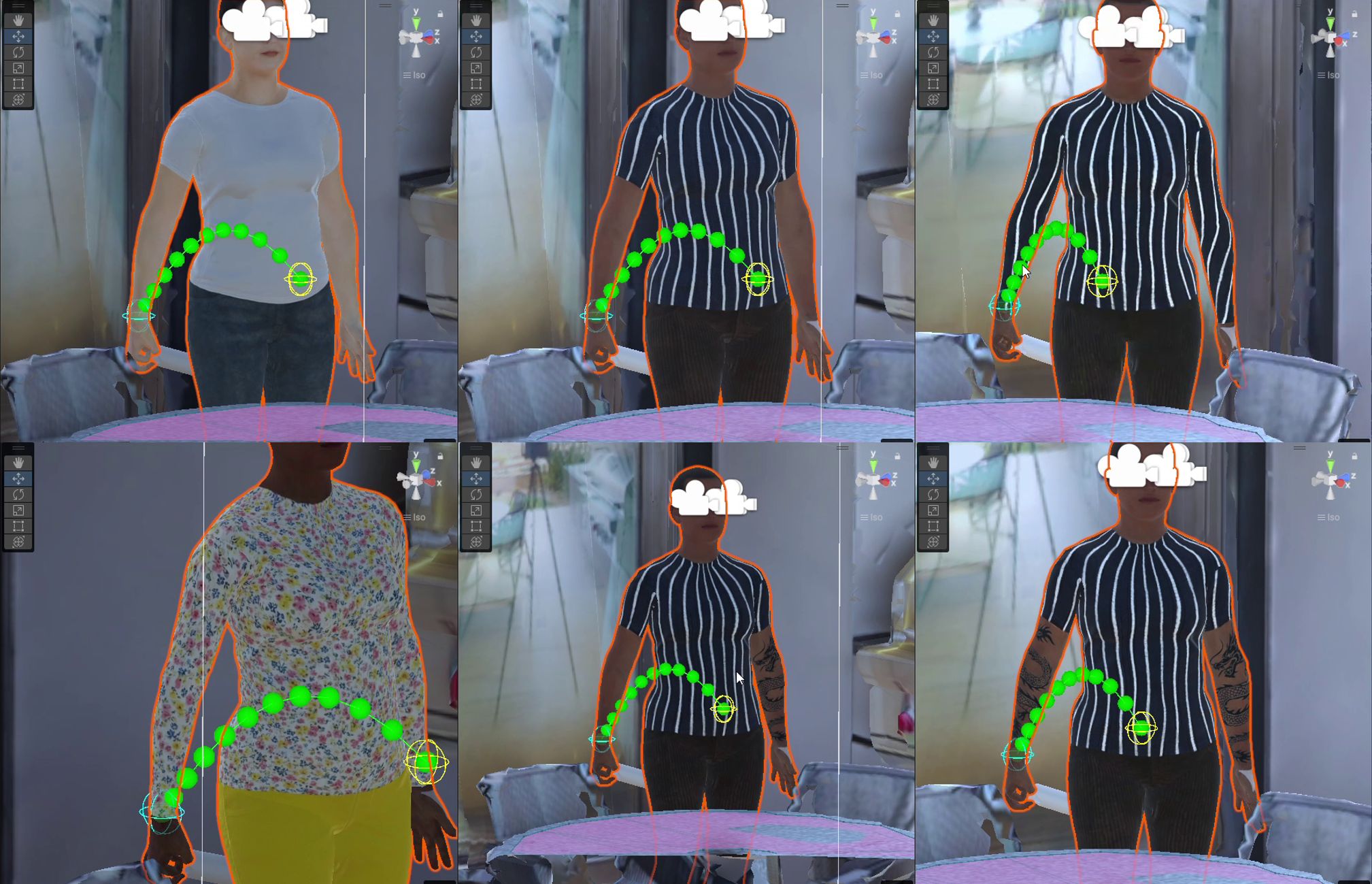}
    \caption{Examples of avatar appearance randomization in \modelname{}. The figure shows variations in clothing texture and visual appearance across episodes.}
    \label{fig:agent_cloth}
\end{figure}

To increase diversity across episodes, \modelname{} applies agent-level randomization at initialization time. In particular, the physical properties of the humanoid agent, such as height and weight, can be varied across episodes to produce a broader range of embodied configurations. In addition to these physical variations, the simulator supports visual randomization of the avatar appearance, including changes in clothing textures and other appearance attributes (see Fig.~\ref{fig:agent_cloth}).

\begin{figure}
    \centering
    \includegraphics[width=0.75\linewidth]{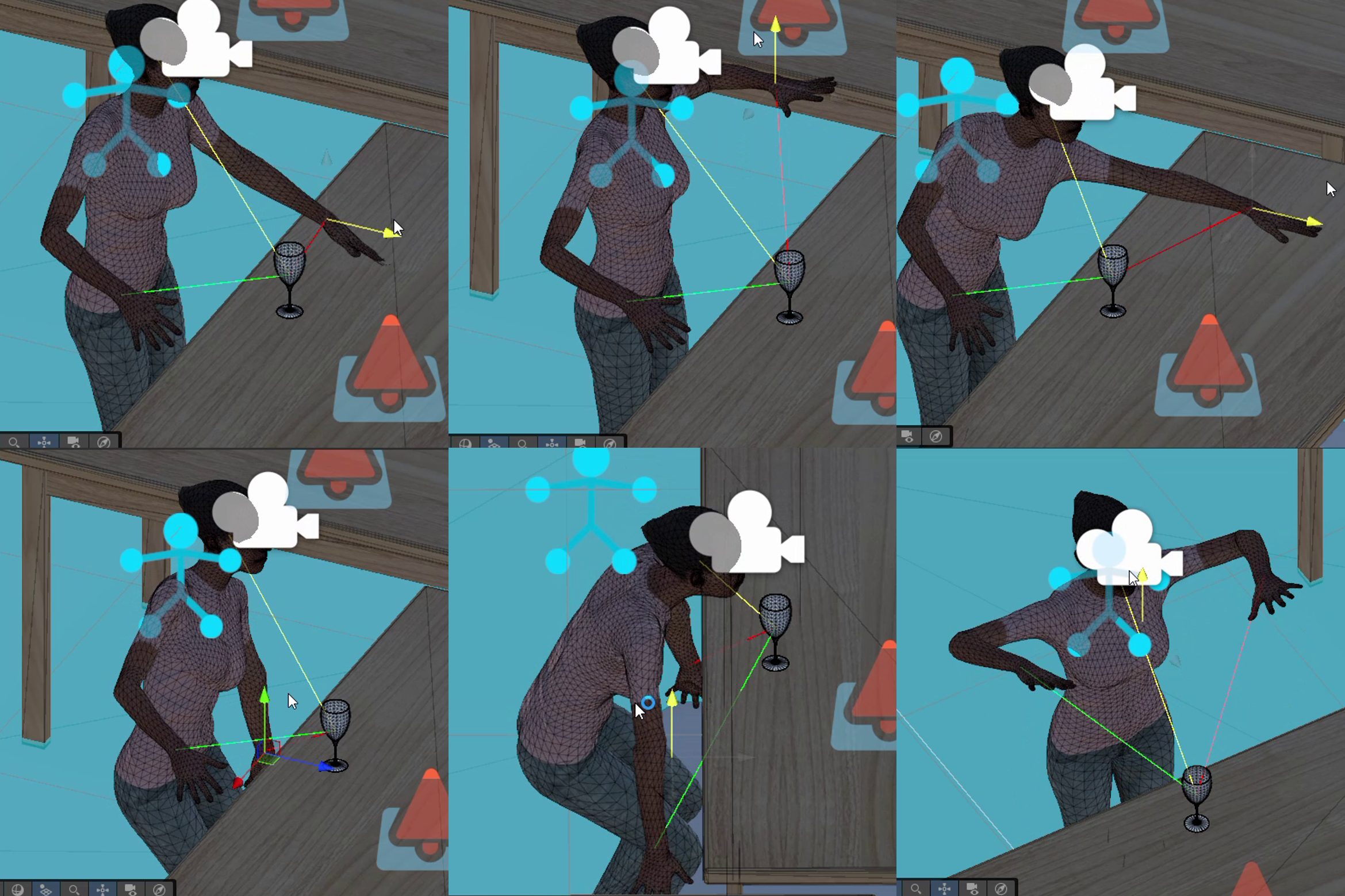}
    \caption{Examples of full-body inverse kinematics in \modelname{}. Given target hand poses, the IK solver generates coordinated whole-body configurations involving the arm, torso, and head to produce plausible egocentric interaction poses.}
    \label{fig:fullbody_ik}
\end{figure}

\paragraph{Pose Generation}
To execute grasping and manipulation actions, \modelname{} maps target hand poses to whole-body humanoid configurations using a full-body inverse kinematics solver. The solver operates over a structured kinematic chain that includes the root, pelvis, spine, head, arms, and legs, allowing the agent to coordinate upper-body reaching with consistent whole-body posture. This produces plausible agent poses in the scene, as illustrated in Fig.~\ref{fig:fullbody_ik}.

\paragraph{Grasp Validation}
To support grasp generation and trajectory validation, \modelname{} uses a simplified collision proxy of the hand. Specifically, the fingers are approximated with capsule colliders and the palm with a box collider, yielding an efficient representation for collision-aware interaction planning. Rather than using a fixed hand proxy, these colliders are adapted to the current avatar by analyzing its mesh, so that the proxy remains consistent with the instantiated humanoid geometry.

\begin{figure}
    \centering
    \includegraphics[width=1\linewidth]{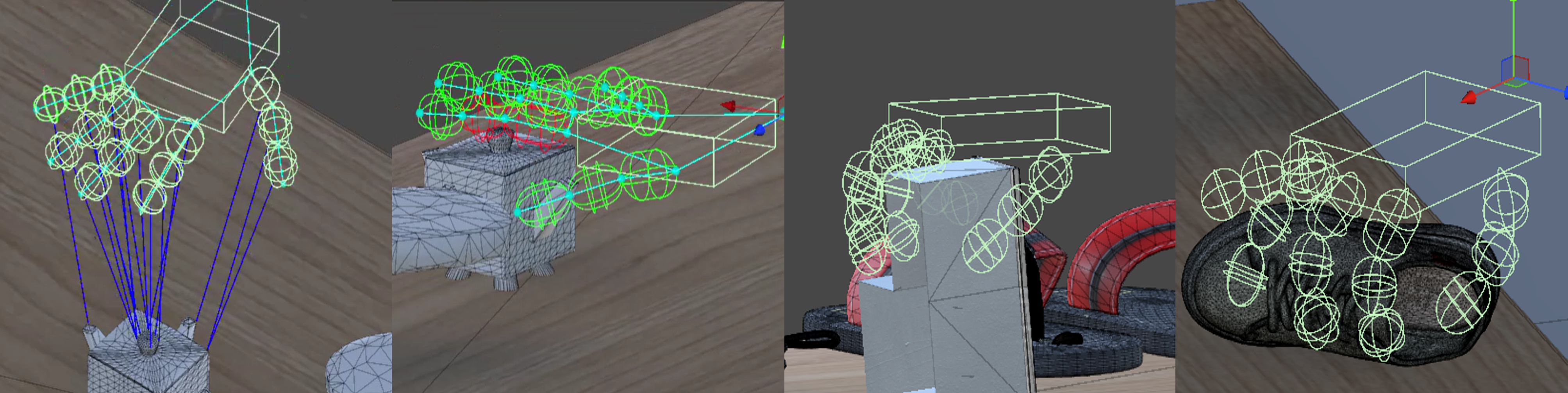}
    \caption{Examples of the collision hand proxy used for grasp generation in \modelname{}. The hand is approximated with capsule colliders for the fingers and a box collider for the palm.}
    \label{fig:grasp_system}
\end{figure}

The resulting hand proxy is used not only to validate the final grasp pose, but also to check intermediate waypoints along the approach trajectory. Collision tests are performed over the full approach sequence against both the environment and distractor objects, allowing invalid interaction trajectories to be rejected before execution. Examples of the proxy representation are shown in Fig.~\ref{fig:grasp_system}.

\paragraph{Trajectory Generation}
To generate the reaching motion, \modelname{} constructs a hand trajectory between the initial hand position and the target grasp position using a Bézier curve. In our implementation, we use 10 intermediate Bézier samples, corresponding to the \texttt{Ultra} sampling setting, together with a control-point height offset of $0.2$ and a control-point exponent of $1.5$. Intermediate waypoints are sampled along the curve and subsequently converted into full-body poses through inverse kinematics.

\begin{figure}
    \centering
    \includegraphics[width=1\linewidth]{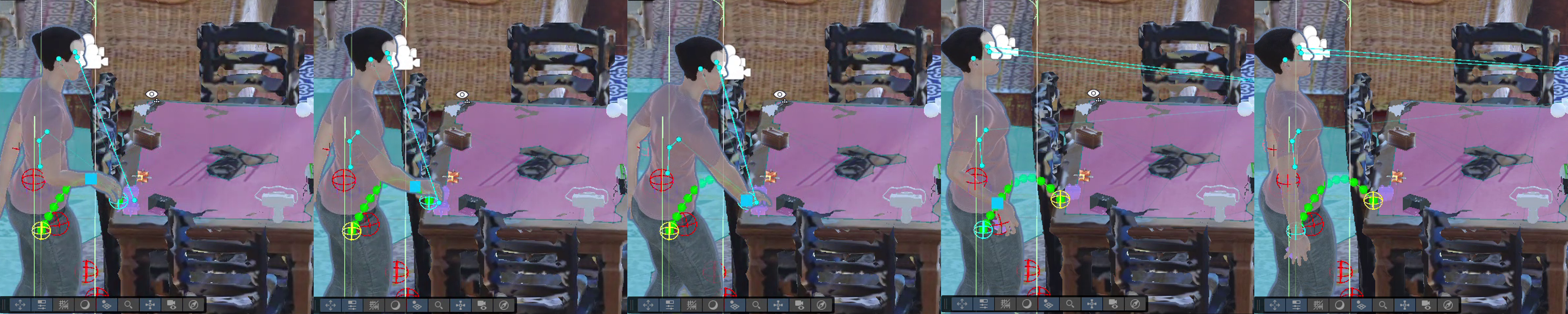}
    \caption{Sequence of full-body poses generated during interaction execution in \modelname{}. The solver coordinates arm, torso, and body motion over time to follow the planned trajectory while maintaining a plausible embodied configuration.}
    \label{fig:grasp_trajectory}
\end{figure}

The control points are obtained by interpolating between the start and target positions and adding a vertical offset along the upward direction. This offset follows a sinusoidal profile, producing an arched approach trajectory rather than a straight-line motion, while the exponent parameter biases the distribution of control points along the path. Candidate trajectories are then validated through collision checks over the full approach sequence before interaction execution. An example of the resulting trajectory is shown in Fig.~\ref{fig:grasp_trajectory}.

\begin{figure}
    \centering
    \includegraphics[width=1\linewidth]{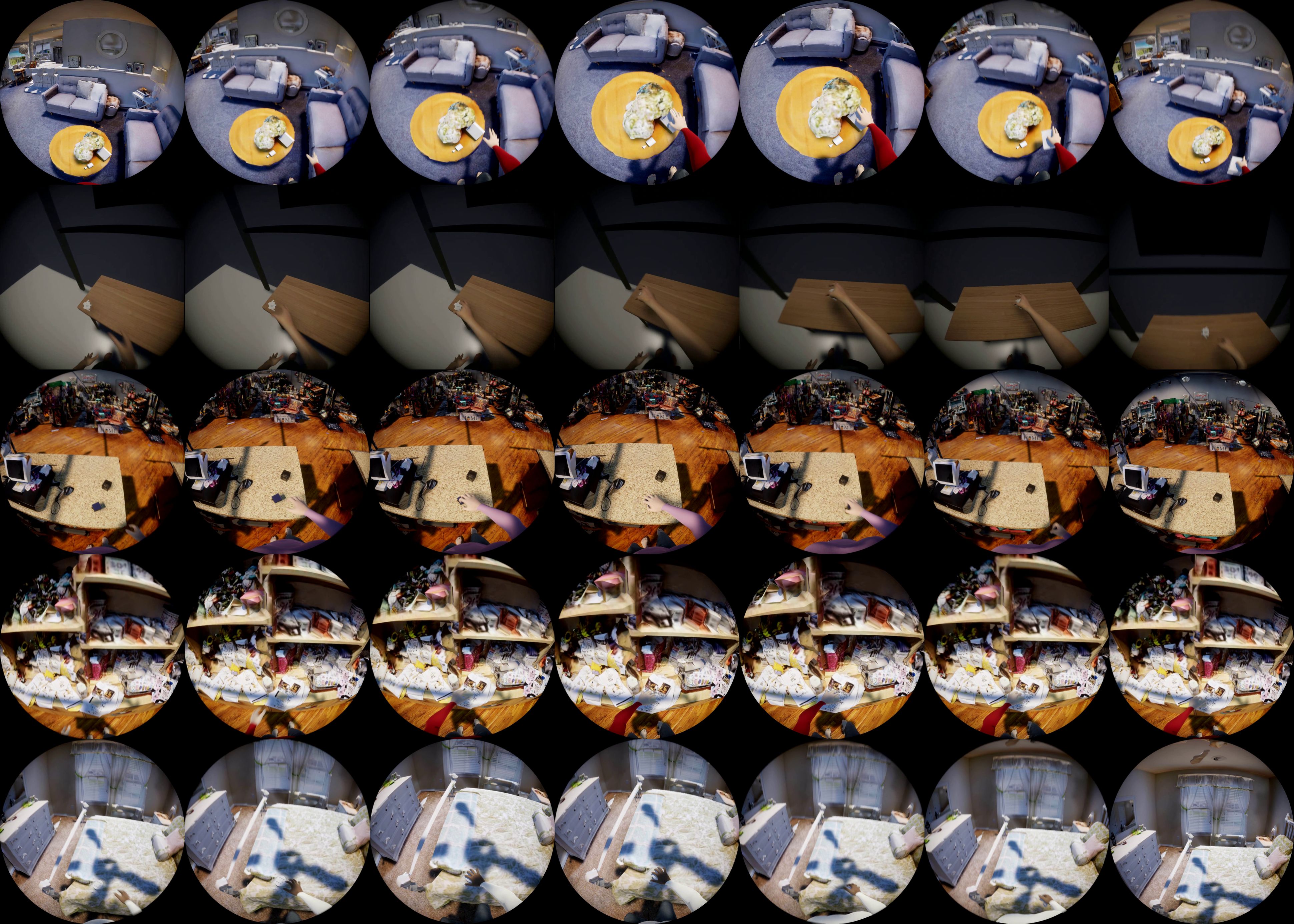}
    \caption{Examples of interaction episodes generated by \modelname{}. Each row shows a different procedurally generated episode in a distinct indoor environment, showing the diversity of egocentric views, object configurations, and scene layouts produced by the simulator.}
    \label{fig:generated_episodes}
\end{figure}

\begin{figure}
    \centering
    \includegraphics[width=1\linewidth]{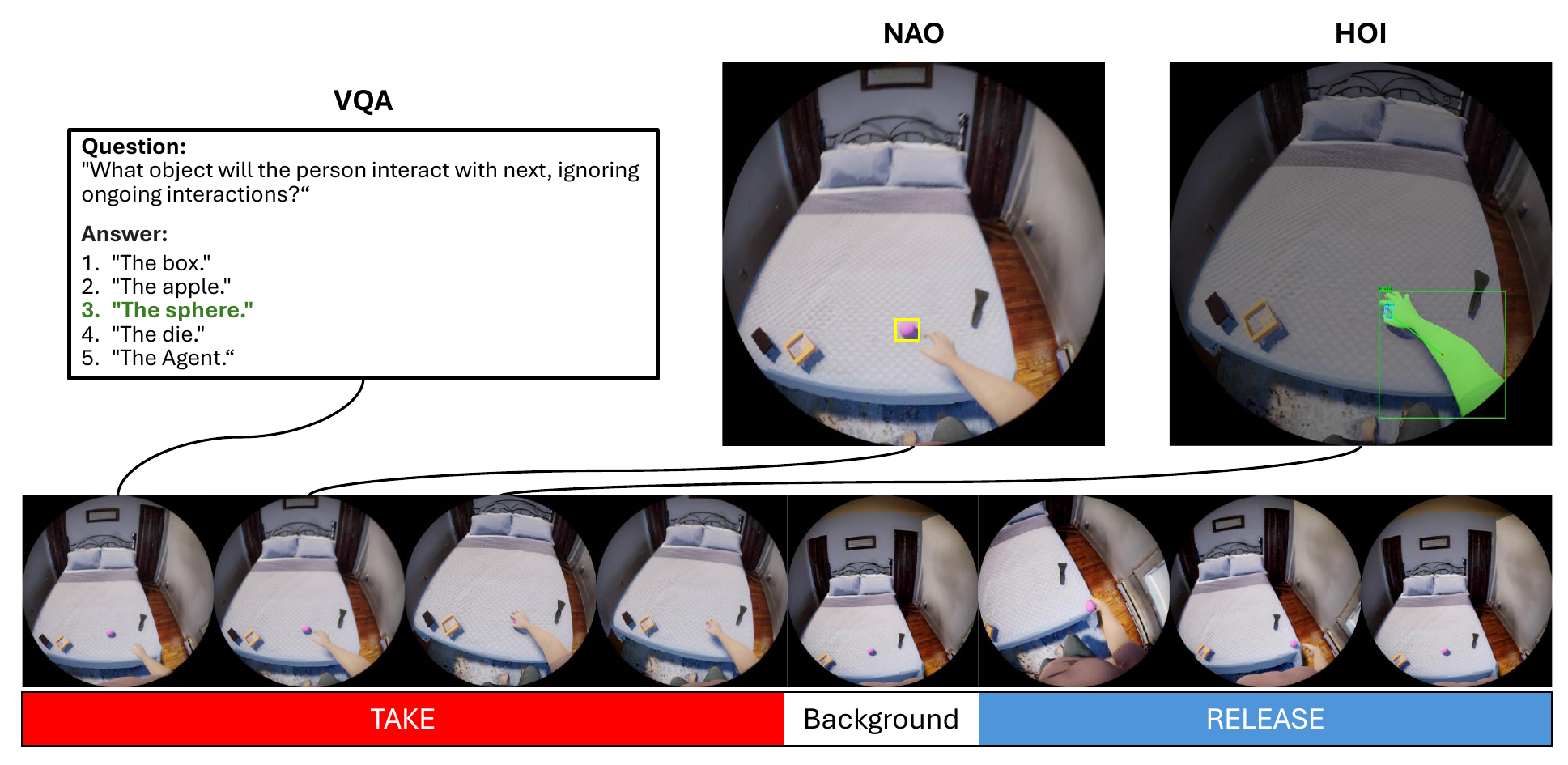}
    \caption{Example of interaction episodes generated by \modelname{} with the set of temporal and spatial annotations obtained automatically.}
    \label{fig:labels}
\end{figure}

\paragraph{Episodes}
Figure~\ref{fig:generated_episodes} shows representative examples of the resulting episodes, highlighting the diversity of indoor scenes, object layouts, and egocentric views produced by the simulator. Figure \ref{fig:labels} shows the set of labels automatically generated for each episode with \modelname{}.

\begin{figure}
    \centering
    \includegraphics[width=1\linewidth]{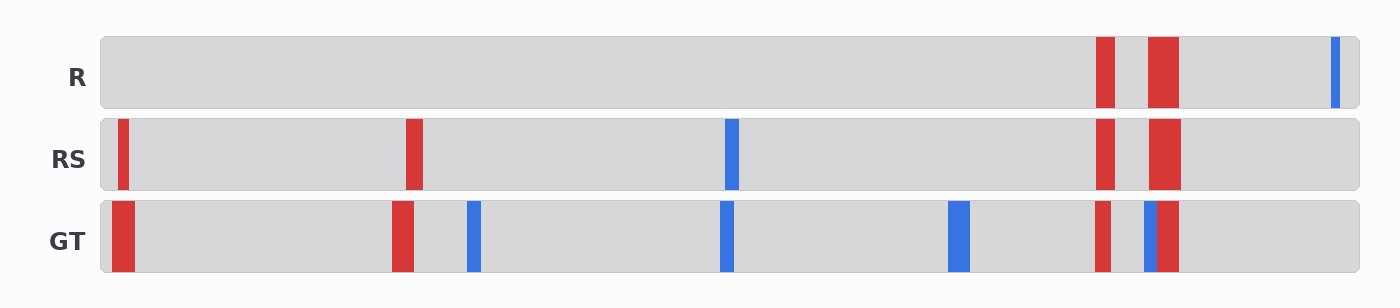}
    
    \vspace{0.3cm}
    
    \includegraphics[width=1\linewidth]{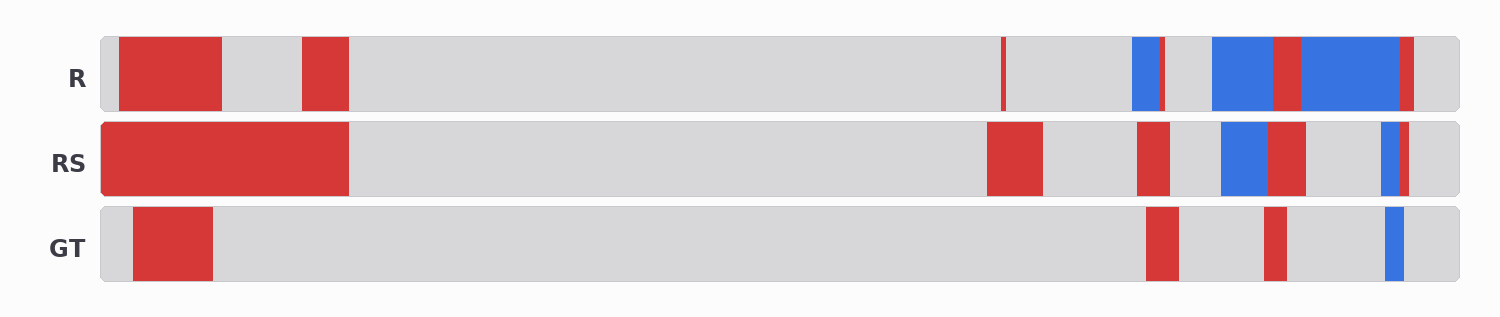}
    
    \caption{Qualitative TAS predictions on representative test videos from \emph{EPIC-KITCHENS} (top) and \emph{Ego-Exo4D} (bottom). Red segments indicate \emph{Take} actions, while blue segments indicate \emph{Release} actions.}
    \label{fig:qualitative_tas}
\end{figure}

\subsection{Experiments and Results}\label{sec_apx_experimental_setup}
In this section, we report additional experimental results, including ablation studies and qualitative examples.

\subsubsection{Temporal Action Segmentation}\label{sec_apx_tas}
\paragraph{Qualitative Results}
To further analyze the impact of synthetic data, we qualitatively evaluate the models trained with 25\% real data on EPIC-KITCHENS and Ego-Exo4D, 
the configuration that yielded the largest performance gap between R and RS. Figure ~\ref{fig:qualitative_tas} shows the frame-wise prediction of both models on representative test video segments, highlighting only \emph{Take}(red) and \emph{Release}(blue) actions against the ground truth.
On EPIC-KITCHENS, the RS model successfully detects three additional Take/Release events that R completely misses, demonstrating a greater capacity of the model.Furthermore, R produces a spurious Release prediction near the end of the sequence that has no correspondence in the ground truth, whereas RS avoids this false positive. On Ego-Exo4D, RS produces predictions that more closely follow the ground truth structure, particularly in the second half of the sequence, while R tends to either over-segment or misplace actions.
In both cases, despite being trained for the same number of epochs, RS exhibits a clear advantage over R, suggesting that synthetic data provides a meaningful learning signal that the real data alone cannot supply at this supervision level.

\subsubsection{Egocentric Hand-Object Interaction Detection}\label{sec_apx_hoi}
\begin{figure}
    \centering
    \includegraphics[width=0.75\linewidth]{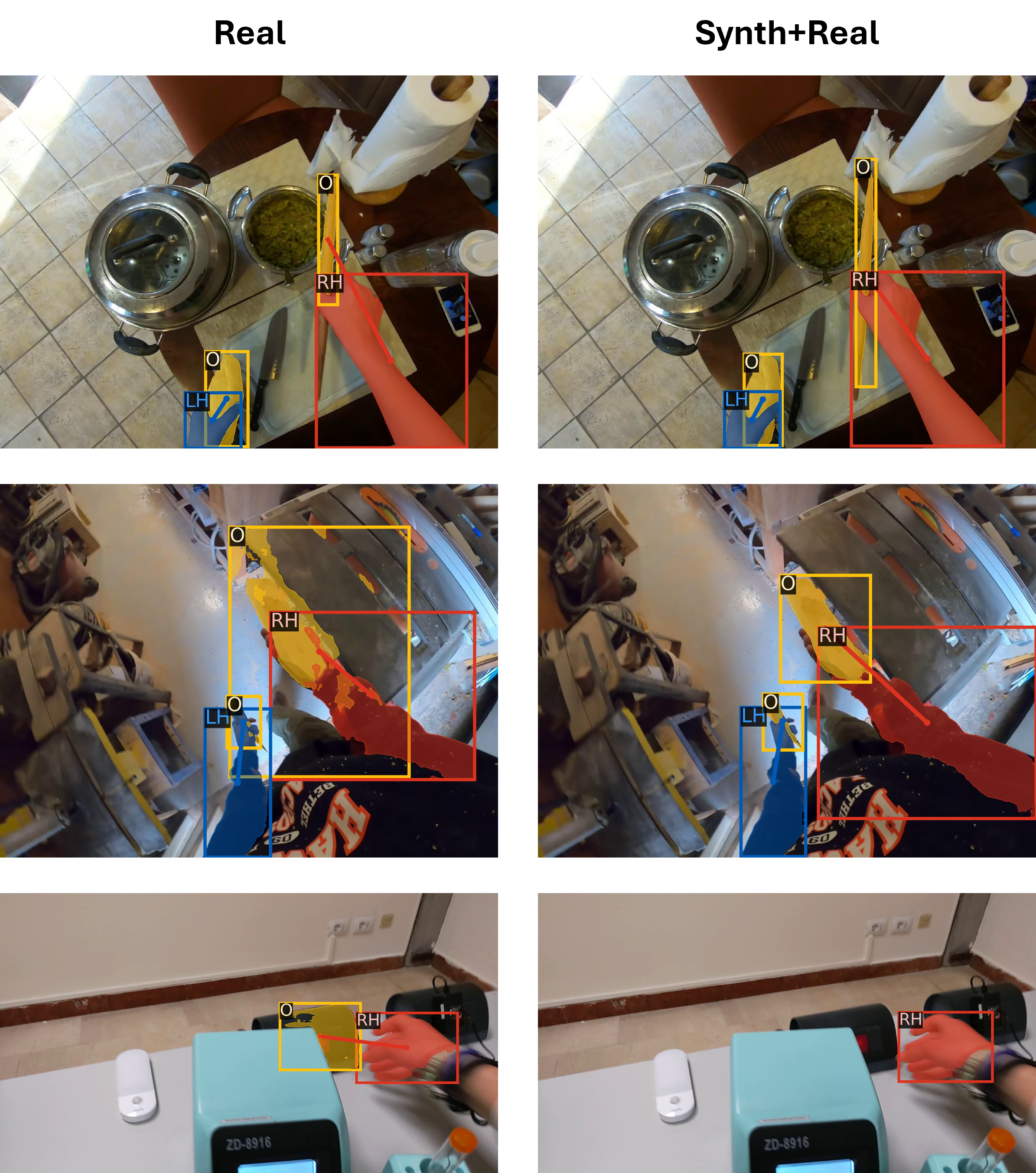}
    \caption{Qualitative comparison between models trained with real data only (\textbf{Real}) and with synthetic plus real data (\textbf{Synth+Real}) on egocentric hand-object interaction segmentation. Adding synthetic data improves the localization of hands and objects and yields more accurate hand-object associations in challenging interaction scenarios. The qualitative results are related to the VISOR, EgoHOS and ENIGMA-51 datasets, respectively.}
    \label{fig:hoi_qualitative}
\end{figure}

\paragraph{Qualitative Results}
Figure~\ref{fig:hoi_qualitative} shows qualitative comparisons between models trained with real data only and with synthetic plus real data. Across the reported examples, augmenting the training set with synthetic samples leads to more accurate hand and object localization and to more reliable hand-object associations. The benefit is especially visible in difficult scenes characterized by occlusions, clutter, and difficult objects.

\subsection{Next-Active Object Detection}
\begin{figure}
    \centering
    \includegraphics[width=0.75\linewidth]{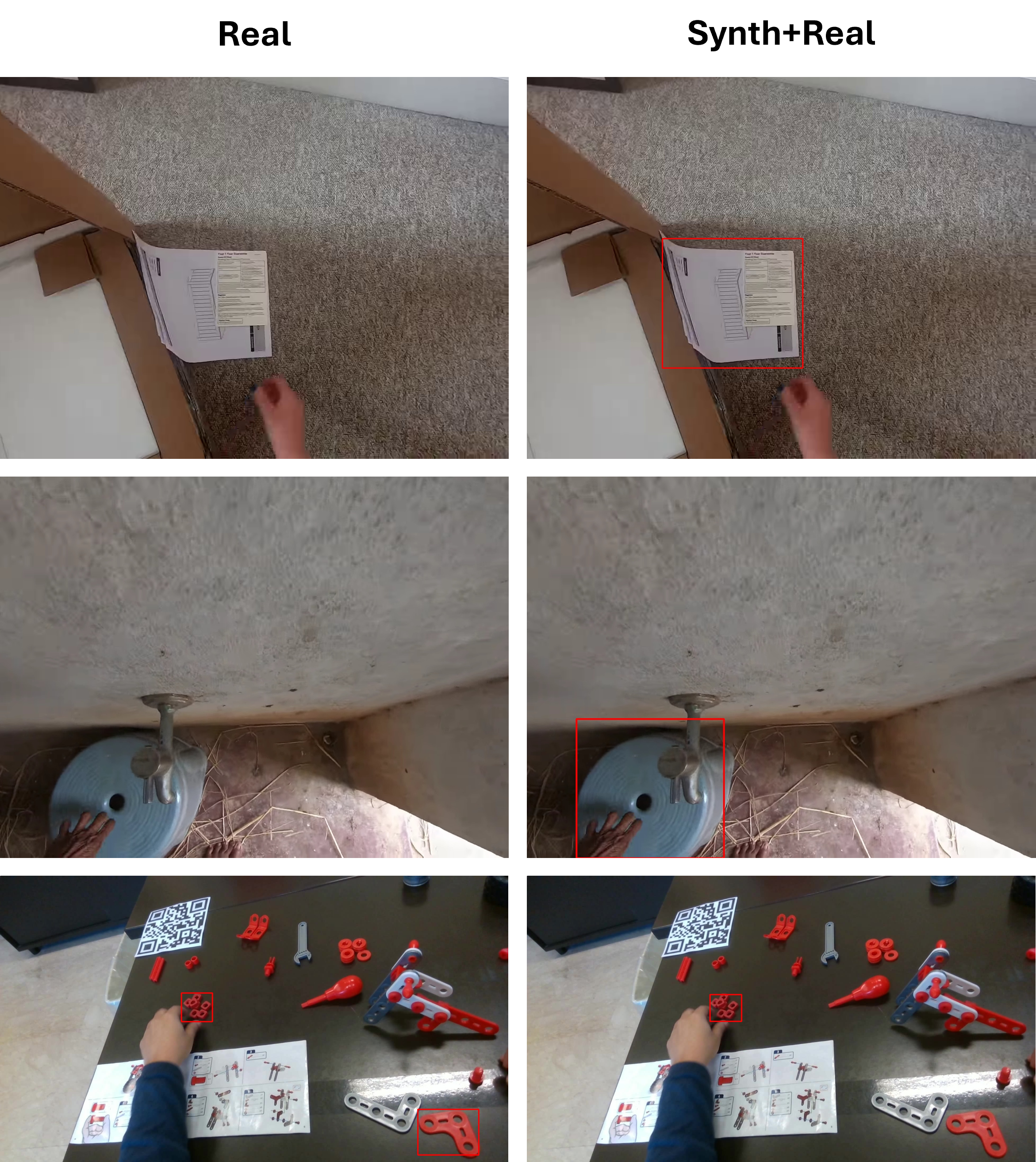}
    \caption{Qualitative comparison between models trained with real data only (\textbf{Real}) and with synthetic and real data (\textbf{Synth+Real}) on next active object (NAO) prediction. Adding synthetic data improves the localization of the future active object and yields more accurate predictions in challenging egocentric scenes.}
    \label{fig:nao_qualitative}
\end{figure}

\paragraph{Qualitative Results}
Figure~\ref{fig:nao_qualitative} presents qualitative comparisons for next active object prediction between models trained with real data only and with synthetic plus real data. Across all examples, synthetic augmentation leads to more accurate localization of the object that will become active in the near future. In the first two cases, the model trained on real data only fails to detect the target object, while the model trained with synthetic augmentation correctly localizes it. In the third example, the real-only model predicts an incorrect additional object, whereas synthetic augmentation leads to the correct identification of the target object only.

\subsubsection{Interaction Anticipation}

\begin{figure}[t]
    \centering
    \begin{minipage}[t]{0.40\linewidth}
        \centering
        \includegraphics[width=\linewidth]{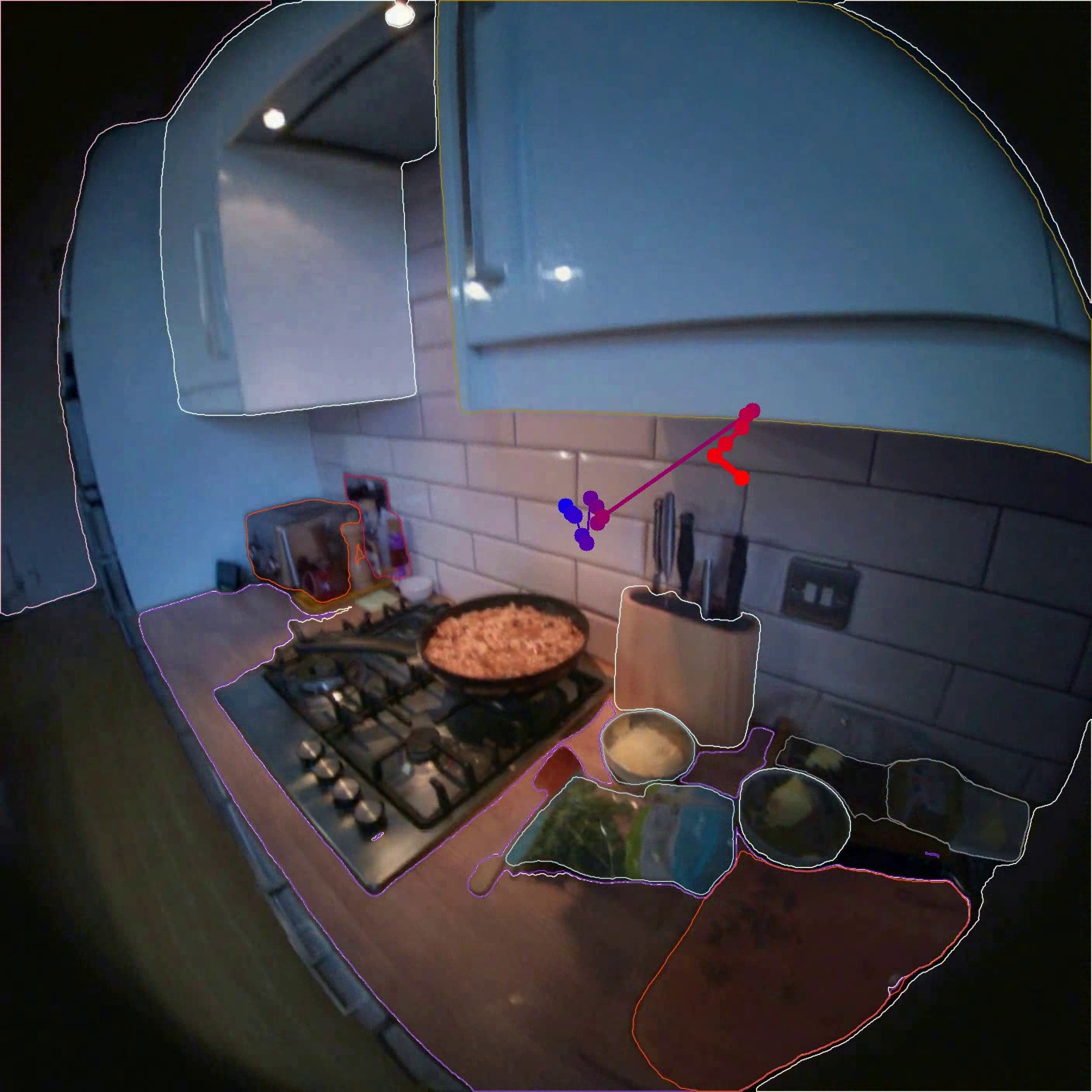}\\[0.3em]
        \raggedright
        \scriptsize
        \textbf{B:} \textcolor{red}{The cheese grater}\\
        \textbf{F:} \textcolor{darkgreen}{The wooden spatula}\\
        \textbf{C:} \textit{The wooden spatula}
    \end{minipage}
    \hfill
    \begin{minipage}[t]{0.40\linewidth}
        \centering
        \includegraphics[width=\linewidth]{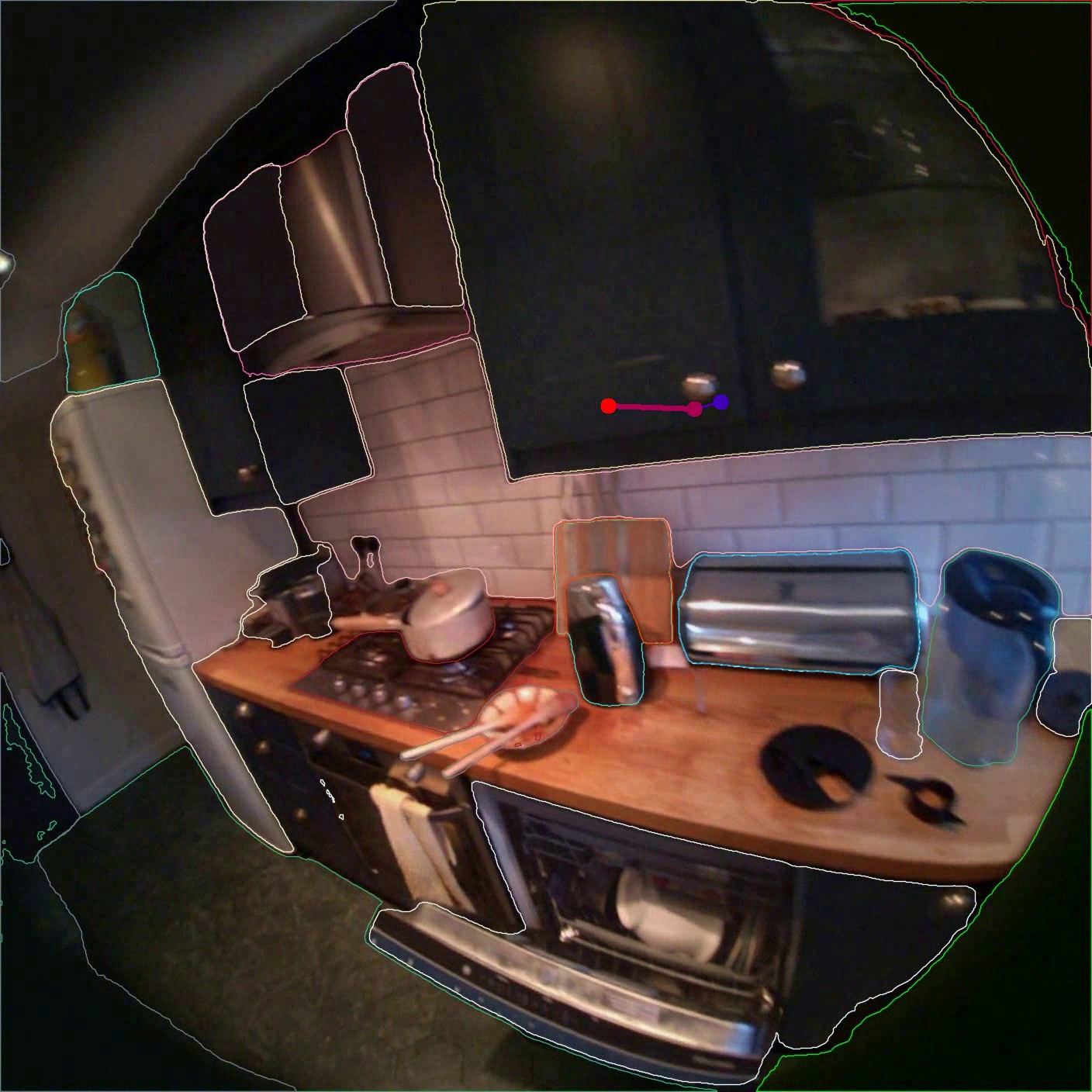}\\[0.3em]
        \raggedright
        \scriptsize
        \textbf{B:} \textcolor{red}{The blue cloth}\\
        \textbf{F:} \textcolor{darkgreen}{The wooden spoon}\\
        \textbf{C:} \textit{The wooden spoon}
    \end{minipage}
    \vspace{1em}

    \begin{minipage}[t]{0.40\linewidth}
        \centering
        \includegraphics[width=\linewidth]{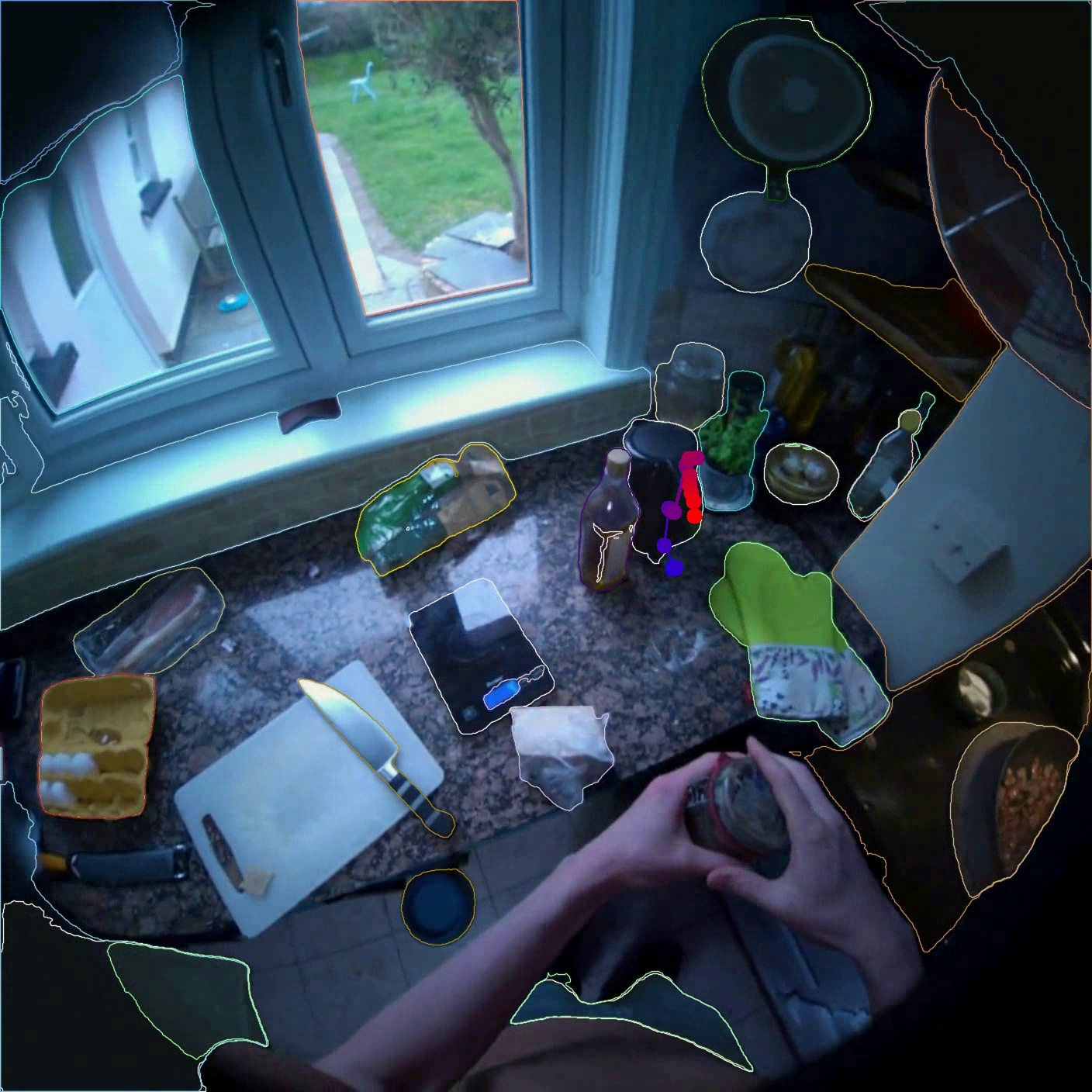}\\[0.3em]
        \raggedright
        \scriptsize
        \textbf{B:} \textcolor{red}{The pot of pepper}\\
        \textbf{F:} \textcolor{darkgreen}{The cheese block}\\
        \textbf{C:} \textit{The cheese block}
    \end{minipage}
    \hfill
    \begin{minipage}[t]{0.40\linewidth}
        \centering
        \includegraphics[width=\linewidth]{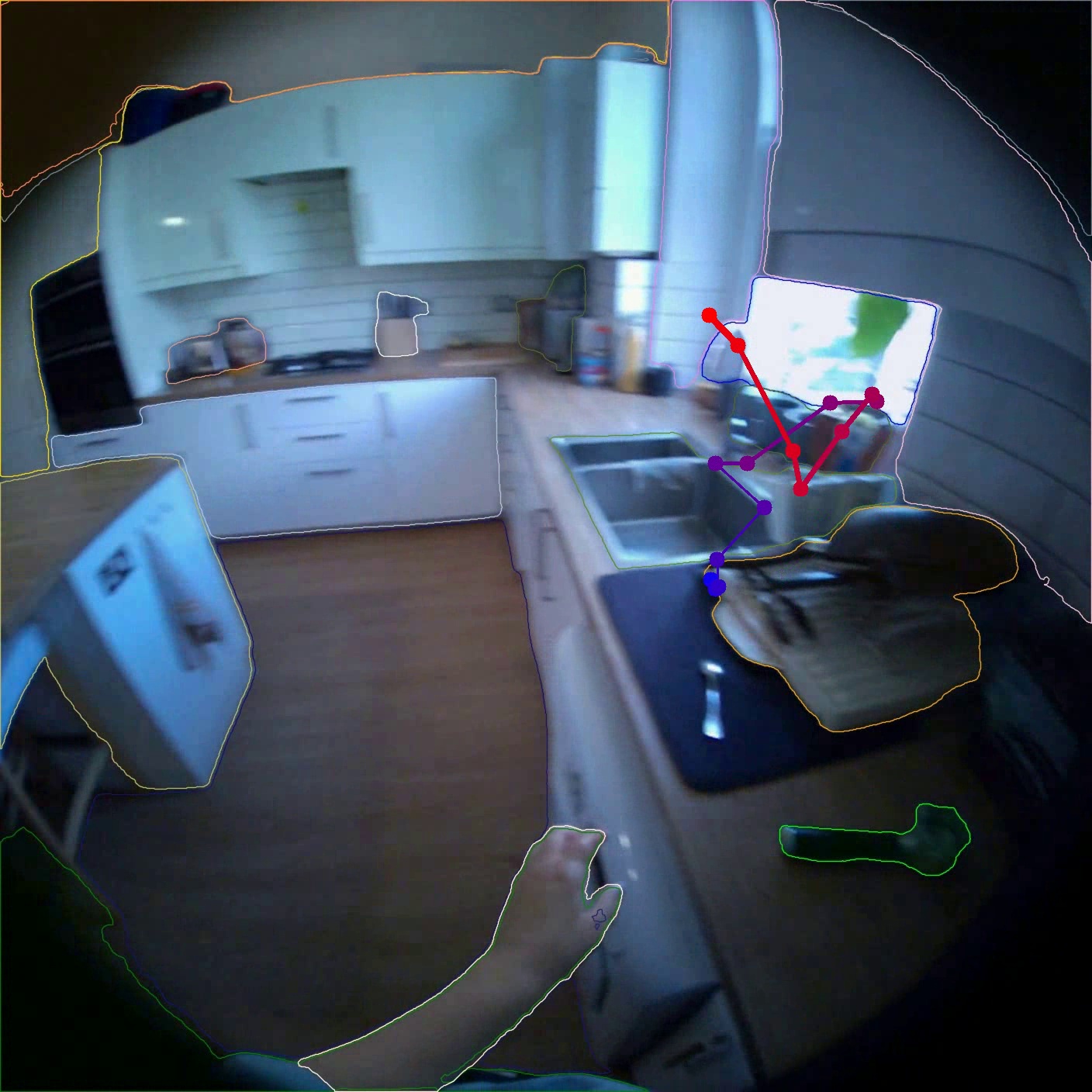}\\[0.3em]
        \raggedright
        \scriptsize
        \textbf{B:} \textcolor{red}{The coffee machine}\\
        \textbf{F:} \textcolor{darkgreen}{The portafilter}\\
        \textbf{C:} \textit{The portafilter}
    \end{minipage}
    \vspace{0.5em}
    \caption{Qualitative examples of synthetic finetuning improving
              interaction anticipation. For each example, the baseline prediction (\textbf{B}, \textcolor{red}{red}) is incorrect, while the finetuned model (\textbf{F}, \textcolor{darkgreen}{green}) correctly 
              identifies the target object (\textbf{C}).}
    \label{fig:qualitative_interaction_anticipation}
\end{figure}

\paragraph{Qualitative Results}
Figure~\ref{fig:qualitative_interaction_anticipation} provides qualitative examples illustrating the improvements yielded by our synthetic finetuning. As observed in the baseline predictions, the original models frequently struggle in cluttered egocentric environments, often misidentifying the target by selecting distractor or adjacent items, especially when visual cues such as the user's gaze are uninformative. In contrast, the finetuned models demonstrate a more robust understanding of the scene.

\subsubsection{Hardware Details}
Our experiments were conducted using two different hardware configurations. For Temporal Action Segmentation, we trained and evaluated models using 4 NVIDIA A30 GPUs with 24 GB of memory each. For Hand-Object Interaction, Next-Active Object, and Interaction Anticipation tasks, we used 4 NVIDIA L40 GPUs with 46 GB of memory each.


\end{document}